\documentclass[11pt]{article}

\PassOptionsToPackage{table}{xcolor}
\usepackage[final]{acl}
\usepackage{times}
\usepackage{latexsym}
\usepackage[T1]{fontenc}
\usepackage[utf8]{inputenc}
\usepackage{microtype}
\usepackage{inconsolata}
\usepackage{graphicx}
\usepackage{amsmath}
\usepackage{amssymb}
\usepackage{mathtools}
\usepackage{amsthm}
\usepackage{algorithm}
\usepackage{algorithmic}
\usepackage[capitalize,noabbrev]{cleveref}
\usepackage{multirow}
\usepackage{adjustbox}
\usepackage{booktabs}
\usepackage{color}
\usepackage{hyperref}

\title{GRKV: Global Regression for Training-Free KV Cache Compression in Long-Context LLMs}

\author{
 \textbf{Junjie Peng\textsuperscript{1,2,3}},
 \textbf{You Wu\textsuperscript{2,4}},
 \textbf{Haoyi Wu\textsuperscript{2,4}},
 \textbf{Jialong Han\textsuperscript{2,4}},
\\
 \textbf{Xiaohua Xie\textsuperscript{1,3}},
 \textbf{Kewei Tu\textsuperscript{2,4}\thanks{Corresponding authors.}},
 \textbf{Jianhuang Lai\textsuperscript{1,3}\footnotemark[1]}
\\
\\
 \textsuperscript{1}School of Computer Science and Engineering, Sun Yat-sen University
\\
 \textsuperscript{2}School of Information Science and Technology, ShanghaiTech University
\\
 \textsuperscript{3}Pazhou Lab (Huangpu)
\\
 \textsuperscript{4}Shanghai Engineering Research Center of Intelligent Vision and Imaging
\\
 \texttt{pengjunjie.pjj2003@gmail.com}, \texttt{tukw@shanghaitech.edu.cn},
\\
 \texttt{stsljh@mail.sysu.edu.cn}
}

\begin{document}
\maketitle
\begin{abstract}
Large language models (LLMs) with extended context lengths rely on the key-value (KV) cache to support attention over prior tokens. However, maintaining the KV cache incurs substantial memory overhead, motivating KV-cache compression methods that enforce a fixed budget through eviction and merging. Modern eviction methods increasingly adopt span-based retention because preserving contiguous spans is empirically effective and better preserves semantic coherence. Yet, when combined with post-eviction merging, span-based retention concentrates merges onto a small set of span-boundary carrier tokens, producing a highly imbalanced merge pattern that exacerbates over-merging and increases information loss. To address this imbalance, we propose \textbf{GRKV} (\textbf{G}lobal \textbf{R}egression for \textbf{KV} Cache), a training-free KV-cache merging method that directly minimizes the discrepancy between compressed-cache and full-cache attention outputs. GRKV uses ridge-regression-based merge steps to distribute information from evicted tokens across retained tokens, while regularizing the updates to prevent over-smoothing. Across the LongBench and RULER long-context benchmarks, GRKV is the only merging method that improves overall performance with minimal overhead. Our code is available at \url{https://github.com/pjunjie/GRKV}.
\end{abstract}

\begin{figure}[t]
\centering
\includegraphics[width=\columnwidth]{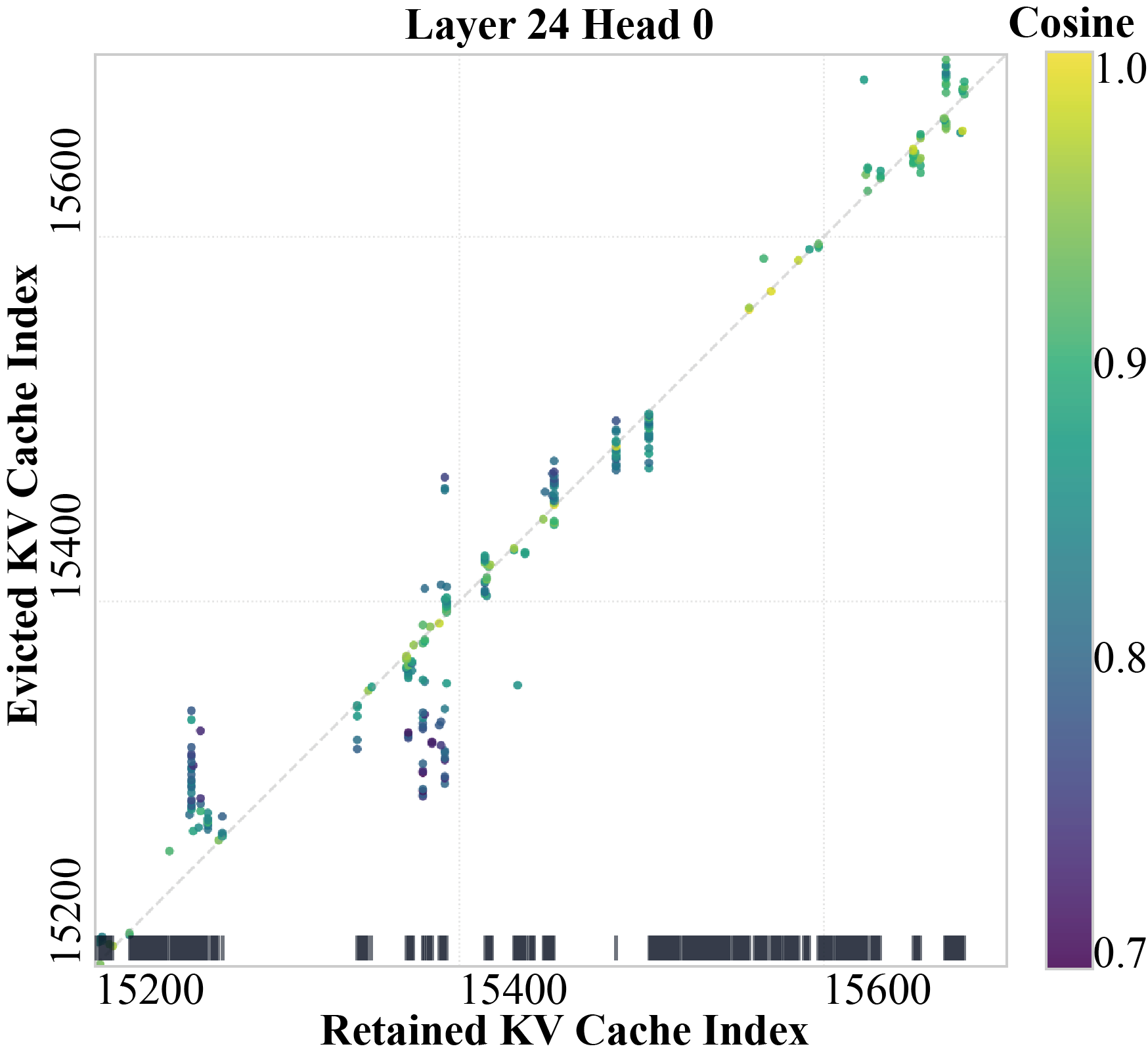}
\caption{\textbf{KV merge map on \textsc{NarrativeQA}.} Using the same key-similarity-based matching strategy as D2O/KVMerger, we merge 250 evicted tokens into the 244 tokens retained by SnapKV. Each point represents a matched (evicted, retained) token pair, and the color indicates the cosine similarity between their keys. Black bars mark the retained-token spans along the \textit{x-axis}.}
\label{fig:kv_merge_map}
\end{figure}

\begin{figure*}[t]
\centering
\includegraphics[width=\textwidth]{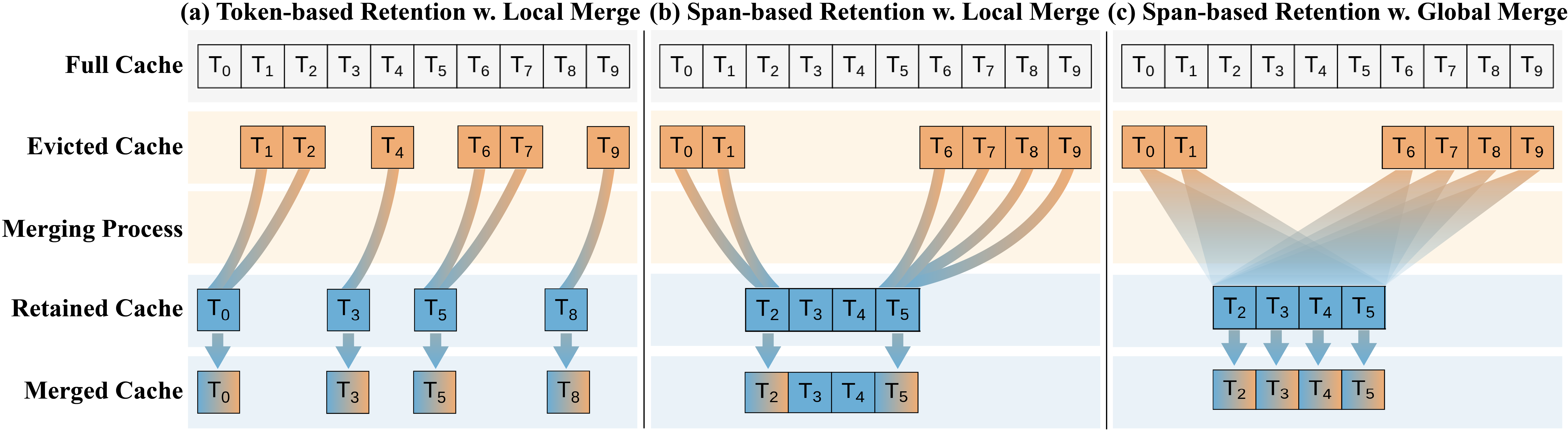}
\caption{\textbf{Overview of how eviction granularity reshapes merge assignments.} \textit{(a)} Token-based retention with a local merge rule, in which evicted tokens are merged into nearby retained tokens, producing relatively dispersed assignments. \textit{(b)} Span-based retention with the same local merge rule, which concentrates many evicted tokens onto a small set of boundary carrier tokens. \textit{(c)} Span-based retention with the global merge rule of GRKV, which uses all retained tokens as carriers and distributes information from the evicted tokens across them.}
\label{fig:overview}
\end{figure*}

\section{Introduction}

Large language models (LLMs)~\citep{achiam2023gpt, touvron2023llama1, touvron2023llama, abdin2024phi, team2024gemma} with extended context lengths rely on the KV cache to support attention over prior tokens. However, maintaining the KV cache incurs substantial memory overhead, motivating a growing body of work on KV-cache compression~\citep{xiao2023efficient, chang2024palu, liu2024minicache, wu-tu-2024-layer, wang2025commonkv}. KV-cache \emph{eviction} methods reduce memory usage by retaining only the most salient tokens---typically those with the highest attention scores---under a fixed cache budget. This inevitably discards some tokens that, while not top-ranked, still carry useful contextual information~\citep{zhang2024cam}. KV-cache \emph{merging} methods have been proposed to recover information from these lower-ranked evicted tokens by incorporating their representations into the retained tokens, without increasing the cache budget.

Early eviction methods selected tokens according to cumulative attention scores~\citep{zhang2023h2o, ge2023model}, which often fragmented contiguous semantic spans and weakened the integrity of the retained text. SnapKV~\citep{li2024snapkv} mitigates this issue by pooling attention scores over segments and then selecting the top-$k$ segments, thereby allowing retained tokens to form contiguous spans that better preserve context. Consequently, SnapKV was among the first methods to adopt span-based retention rather than token-based retention. Many subsequent eviction methods have adopted this span-based strategy~\citep{cai2024pyramidkv, fu2024not, feng2025identify, feng2024ada}.

In contrast, most prior KV-cache merging methods were designed primarily for earlier token-based retention strategies~\citep{zhang2024cam, wang2024model, wan2024d2o, cui2025homogeneous}. Broadly, these methods can be grouped into two classes: (i) local, adjacency-based matching and (ii) key-similarity-based matching. Adjacency-based matching methods (e.g., CaM~\citep{zhang2024cam}, AsymKV~\citep{cui2025homogeneous}) leverage the high similarity between neighboring tokens to merge adjacent tokens using carefully weighted averaging, yielding more compact representations. Key-similarity-based methods (e.g., KVMerger~\citep{wang2024model}, D2O~\citep{wan2024d2o}) merge tokens when their keys are highly similar, based on the observation that keys largely determine attention weights, so similar keys imply similar importance patterns. Prior work, together with our observations in Fig.~\ref{fig:kv_merge_map}, indicates that adjacent tokens often have highly similar keys~\citep{wang2024model}. Thus, key-similarity-based matching can be seen as a generalization of adjacency-based matching over a broader candidate neighborhood.

As eviction methods have increasingly shifted from token-based retention~\citep{zhang2023h2o, liu2023scissorhands, ge2023model, oren-etal-2024-transformers, ren2024efficacy} to span-based retention~\citep{li2024snapkv, cai2024pyramidkv, fu2024not, feng2025identify, feng2024ada}, the merge assignments produced by prior KV-cache merging methods have become markedly more concentrated. In particular, because most merging methods rely on local heuristics, they often funnel many evicted tokens into a small set of span-boundary tokens. These boundary tokens therefore become the main information carriers, overloading their representations and making them prone to over-merging: excessive aggregation can blur or even erase their original semantics, thereby degrading overall performance. A mathematical analysis is provided in Appendix~\ref{appendix:A}. Fig.~\ref{fig:kv_merge_map} provides a concrete example on \textsc{NarrativeQA}~\citep{kocisky-etal-2018-narrativeqa}: we use the same key-similarity-based matching strategy as in D2O/KVMerger to merge evicted tokens into the KV cache retained by SnapKV, and observe that merge assignments concentrate near span boundaries. Additional visualizations in Appendix~\ref{appendix:B} further confirm this concentration pattern. We further quantify this effect across models, benchmarks, cache budgets, and eviction methods under the key-similarity-based matching strategy as in D2O/KVMerger. Boundary carriers constitute only 16--21\% of retained carriers but receive 39--49\% of merge assignments, with 3.09--4.53$\times$ the load of interior carriers. Full definitions and results are provided in Appendix~\ref{sec:carrier_quantitative}.

To address the challenges introduced by span-based retention and illustrated in Fig.~\ref{fig:overview}, we propose \textbf{GRKV} (\textbf{G}lobal \textbf{R}egression for \textbf{KV} Cache), a \emph{global} KV-cache merging method that aligns the compressed cache with the full cache by directly minimizing the discrepancy in attention outputs. We achieve this alignment with ridge-regression-based merge steps that use the full cache as the information source and distribute information from evicted tokens across retained tokens, while regularizing the updates to prevent over-smoothing and carrier-token blurring. In contrast to prior local heuristics that concentrate merges onto a small set of span-boundary tokens, GRKV treats \emph{all} retained tokens as merge carriers, mitigating over-merging by expanding the effective carrier capacity.

Our contributions can be summarized as follows:
\begin{enumerate}
\item We propose GRKV, a training-free KV-cache merging method that explicitly models the discrepancy between the compressed-cache and full-cache attention outputs. By minimizing this discrepancy, GRKV provides a principled objective for optimizing the merging process.
\item GRKV is the first \emph{global} KV-cache merging method designed specifically for modern span-based retention. By enlarging the pool of carrier tokens, GRKV increases the capacity to incorporate information from evicted tokens, thereby better preserving otherwise discarded context. Moreover, GRKV mitigates over-merging through ridge-regression regularization, which curbs carrier-token blurring and reduces information loss.
\item
Across 16 LongBench and 13 RULER tasks, GRKV is the only KV-cache merging method that improves overall performance when combined with modern span-based KV-cache eviction methods, while remaining a plug-and-play module with minimal added overhead.
\end{enumerate}

\section{Related Work}

For token-level KV-cache compression, in which the compressed units are individual KV-cache entries, existing methods can be categorized into \emph{eviction-based} and \emph{merging-based} methods.

\textbf{Eviction-based compression.} \emph{Training-free} eviction methods typically rely on attention-derived importance scores. H2O~\citep{zhang2023h2o} maintains heavy-hitter tokens using accumulated attention scores. SnapKV~\citep{li2024snapkv} identifies important tokens using attention scores computed from a query window. PyramidKV~\citep{cai2024pyramidkv} allocates more KV-cache capacity to lower layers while assigning a smaller budget to higher layers. CriticalKV~\citep{feng2025identify} identifies and retains critical tokens under a perturbation constraint. Ada-KV~\citep{feng2024ada} allocates more budget to heads that model long-range dependencies. \emph{Training-based} eviction methods learn specialized retention or attention patterns. DuoAttention~\citep{xiao2024duoattention} splits attention heads to reduce memory through a specialized training procedure. HeadKV~\citep{fu2024not} performs compression through learned dropping and abstraction.

\textbf{Merging-based compression.} KV-cache merging methods reduce information loss by merging evicted tokens into retained ones instead of simply discarding them. CaM~\citep{zhang2024cam} merges evicted tokens into retained tokens using attention-weighted sampling. KVMerger~\citep{wang2024model} clusters tokens and replaces each cluster with a pseudo-token constructed by weighted averaging. D2O~\citep{wan2024d2o} dynamically chooses between eviction and merging at different token positions. AsymKV~\citep{cui2025homogeneous} applies different local merging strategies for homogeneous keys and heterogeneous values.

\textbf{Comparison to prior work.} Inspired by merge-focused work such as CaM, we study the design of a general merging method for modern span-based KV-cache eviction methods. While GRKV is closely related to prior \emph{merging} methods, it differs by explicitly optimizing a \emph{global} reconstruction objective rather than relying on \emph{local} heuristics. By distributing recovered information across a larger set of retained tokens, GRKV reduces the burden on boundary carrier tokens and more faithfully preserves the attention outputs of the full cache.

\section{Methods}

\subsection{Preliminary}

\textbf{Standard attention computation.} Autoregressive inference in an LLM typically consists of two phases: \emph{prefilling} and \emph{decoding}. For simplicity, we describe a single attention head. In the prefilling phase, the model computes and stores the KV cache for all input tokens, so that the corresponding KV states need not be recomputed during subsequent decoding steps~\citep{ott-etal-2019-fairseq, wolf-etal-2020-transformers}. Concretely, $K = H W^K$ and $V = H W^V$, where $H \in \mathbb{R}^{n \times d}$ denotes the hidden-state matrix for $n$ input tokens, $d$ is the head dimension, and $W^K, W^V \in \mathbb{R}^{d \times d}$ are the key and value projection matrices. In the decoding phase, at the $i$-th decoding step, the model forms a query vector $q_i = H_i W^Q$ with query projection $W^Q \in \mathbb{R}^{d \times d}$. This query is matched against all cached keys to obtain attention weights, which are then used to aggregate the values and produce the output representation: $o_i = A_i V W^O$, where $A_i = \operatorname{softmax}(q_i K^\top / \sqrt{d})$. Here $W^O \in \mathbb{R}^{d\times d}$ denotes the output projection matrix, and $A_i$ is the attention-weight vector over all cached tokens.

\textbf{Full vs.\ retained KV cache.} Let the uncompressed KV cache be denoted by $K_{\text{full}}, V_{\text{full}} \in \mathbb{R}^{n \times d}$. To lower memory consumption, an eviction method retains only $c$ KV-cache entries, producing a reduced cache $K_{\text{Ret}}, V_{\text{Ret}} \in \mathbb{R}^{c \times d}$ with $c < n$. For a query $q_i$, the full-cache attention output is $o_i^{(\text{full})} = A_i^{(\text{full})} V_{\text{full}} W^O$, where $A_i^{(\text{full})} = \operatorname{softmax}(q_i K_{\text{full}}^\top / \sqrt{d})$. Using the retained cache, the output becomes $o_i^{(\text{Ret})} = A_i^{(\text{Ret})} V_{\text{Ret}} W^O$, where $A_i^{(\text{Ret})} = \operatorname{softmax}(q_i K_{\text{Ret}}^\top / \sqrt{d})$. The difference between $o_i^{(\text{full})}$ and $o_i^{(\text{Ret})}$ quantifies the information lost due to compression: a larger gap indicates greater deviation from the full-cache result. This motivates the following per-query discrepancy loss:
\begin{equation}
  \min_{K_{\text{Ret}}, V_{\text{Ret}}} \mathcal{L}_i
  = \,\|\,o_i^{(\text{full})} - o_i^{(\text{Ret})}\,\|_2^2
  = \,\|\,\Delta_i W^O\,\|_2^2~,
\end{equation}
where we define $\Delta_i = A_i^{(\text{full})} V_{\text{full}} - A_i^{(\text{Ret})} V_{\text{Ret}}$ as the difference between the full-cache and retained-cache pre-projection attention outputs for query $i$. Intuitively, $\mathcal{L}_i$ measures how much the compressed cache changes the model's output for that query.

\begin{figure}[t]
\centering
\includegraphics[width=\columnwidth]{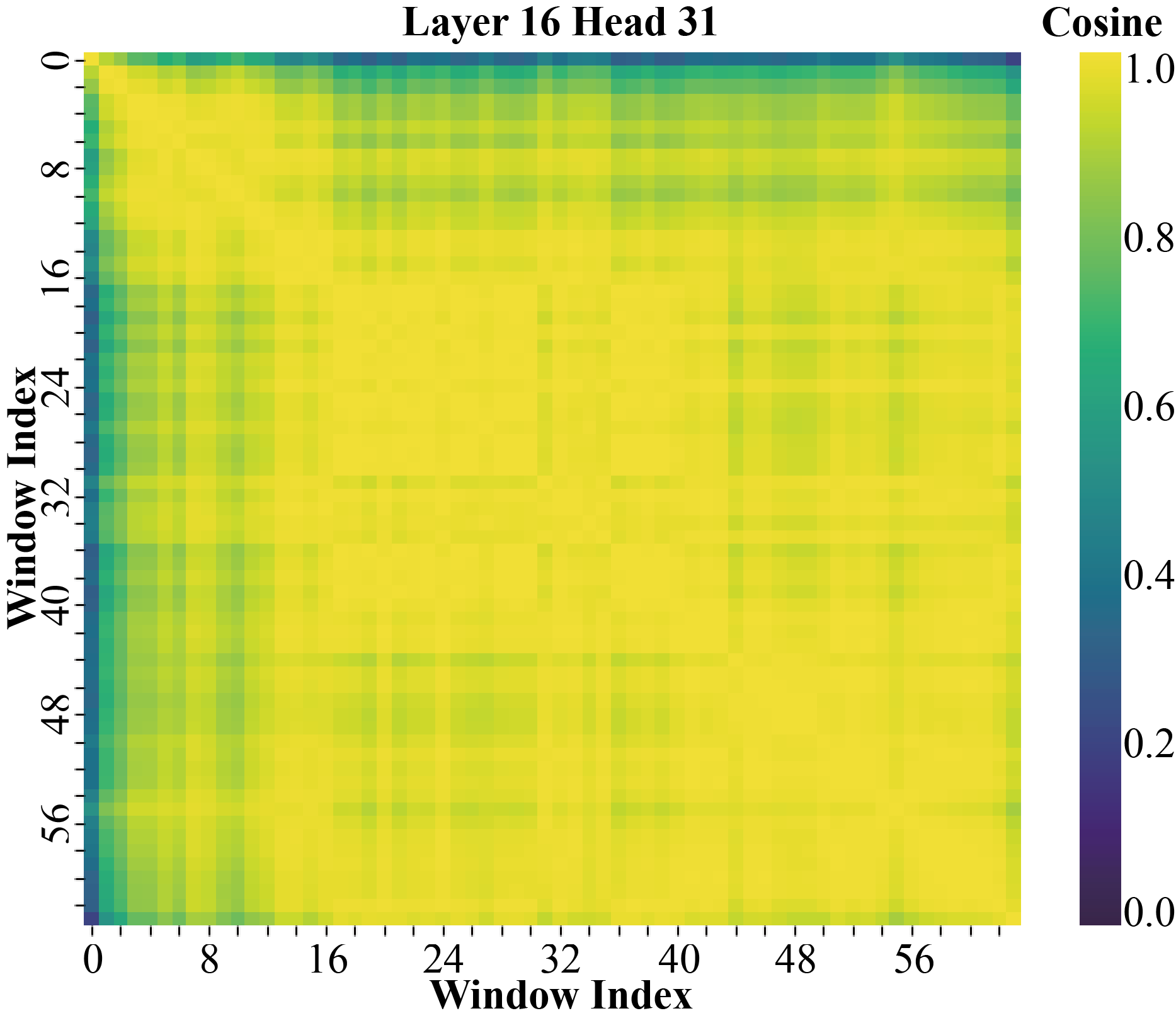}
\caption{\textbf{Cross-window consistency of window-level attention outputs on \textsc{HotpotQA}.} Each cell reports the cosine similarity between two window summaries $s_W$ computed from full-cache attention outputs.}
\label{fig:window_consistency}
\end{figure}

\textbf{Objective decomposition.} To simplify the optimization, we derive an upper bound on $\mathcal{L}_i$ by separating out the output projection $W^O$, which is fixed for a given model. By submultiplicativity of matrix norms, $\|\Delta_i W^O\|_2^2 \,\le\, \|\Delta_i\|_2^2\,\|W^O\|_2^2$. Since $\|W^O\|_2^2$ is constant for a given model, we instead minimize the surrogate objective:
\begin{equation}
\min_{K_{\text{Ret}}, V_{\text{Ret}}} \widetilde{\mathcal{L}}_i = \|\Delta_i\|_2^2~.
\end{equation}
Minimizing $\widetilde{\mathcal{L}}_i$ avoids the additional overhead of explicitly applying $W^O$, while still promoting consistency with the full-cache attention output.

Notably, this objective naturally leverages all retained tokens as a carrier set for absorbing and consolidating information from evicted tokens. In contrast to local merging heuristics that funnel many evicted tokens into a small set of span-boundary carriers---overloading those tokens and increasing information loss---the GRKV objective promotes a broader set of retained tokens as merge targets. This more uniform distribution across retained tokens mitigates over-merging by reducing the imbalance among carrier tokens.

\subsection{Estimating and Optimizing with a Surrogate Query Window}
\label{sec:query_window}

\textbf{Surrogate query window.} During generation, future queries are unknown, so the compressed cache cannot be optimized for each future query individually. GRKV instead uses a short prompt-derived query window as a surrogate. We find that attention outputs computed from different query windows within the same long context are highly consistent, indicating that a late-prompt query window provides a stable proxy for unseen future queries. Specifically, we conduct this analysis on \textsc{HotpotQA}~\citep{yang-etal-2018-hotpotqa}. For each sequence, we treat the first $90\%$ of tokens as the prompt and the remaining $10\%$ as future-query tokens, and further partition the future-query segment into windows of length $m{=}32$. For each window $W$, we compute the mean full-cache attention output:
\begin{equation}
s_W = \frac{1}{m} \sum_{i \in W} A_i^{(\text{full})} V_{\text{full}}.
\end{equation}
We then measure cross-window consistency using the cosine similarity between summary vectors $s_W$ from different windows within the same sequence. As shown in Fig.~\ref{fig:window_consistency}, these window-level summaries remain highly similar across the sequence. Additional visualizations are provided in Appendix~\ref{appendix:C}.

Notably, optimizing with respect to a single query can overfit to its attention pattern, which is undesirable since the query distribution may shift during generation. A multi-query window provides a more stable optimization target. Following~\citet{li2024snapkv}, GRKV uses the final query window of the prompt as its surrogate query window. A position ablation with early, middle, and final query windows yields improvements over the base eviction method, but the variation and the retrieval-oriented failure analysis in Appendix~\ref{sec:surrogate_robustness} show that the surrogate assumption is empirical rather than position-invariant or guaranteed.

\textbf{Window-level objective.} Given a query window $W$ containing $m$ queries, we aggregate the discrepancy over all queries in the window:
\begin{equation}
\min_{K_{\text{Ret}}, V_{\text{Ret}}} \widetilde{\mathcal{L}}_{\text{win}} = \big\|A_{\text{win}}^{(\text{full})} V_{\text{full}} - A_{\text{win}}^{(\text{Ret})} V_{\text{Ret}}\big\|_F^2~,
\end{equation}
where $A_{\text{win}}^{(\text{full})}$ and $A_{\text{win}}^{(\text{Ret})}$ are the attention-weight matrices for $m$ queries in the window. Specifically, $A_{\text{win}}^{(\text{full})} = \operatorname{softmax}(Q_{\text{win}} K_{\text{full}}^\top / \sqrt{d})$ and $A_{\text{win}}^{(\text{Ret})} = \operatorname{softmax}(Q_{\text{win}} K_{\text{Ret}}^\top / \sqrt{d})$, where $Q_{\text{win}} \in \mathbb{R}^{m \times d}$ contains the $m$ queries in the window.

\subsection{Ridge-Regularized KV Reconstruction}

Directly optimizing the window-level objective can cause the retained KV cache to deviate too far from the initial post-eviction cache, leading to unstable changes in attention weights or overly smoothed values. We therefore regularize both $K_{\text{Ret}}$ and $V_{\text{Ret}}$ toward their initial post-eviction cache, $K_{\text{Ret}}^0$ and $V_{\text{Ret}}^0$. The resulting ridge-regularized objective is:
\begin{equation}
\begin{aligned}
\min_{K_{\text{Ret}}, V_{\text{Ret}}} \widetilde{\mathcal{L}}_{\text{KV}} = &\widetilde{\mathcal{L}}_{\text{win}} + \lambda_k\left\|K_{\text{Ret}}-K_{\text{Ret}}^0\right\|_F^2 \\ &+ \lambda_v\left\|V_{\text{Ret}}-V_{\text{Ret}}^0\right\|_F^2~.
\end{aligned}
\label{eq:eq5}
\end{equation}
The coefficients $\lambda_k>0$ and $\lambda_v>0$ control the strength of regularization for keys and values, respectively. Larger coefficients keep the compressed cache closer to its initial post-eviction cache, while smaller coefficients allow more aggressive reconstruction of the full-cache attention outputs.

\subsection{Alternating KV-Cache Optimization}

GRKV alternates between two updates: with the current keys fixed, it solves a ridge-regression problem for $V_{\text{Ret}}$; with the current values fixed, it applies a local linearized ridge update to $K_{\text{Ret}}$.

\textbf{Value step.} With the current keys $K_{\text{Ret}}$ fixed, we update $V_{\text{Ret}}$ by minimizing:
\begin{equation}
\min_{V_{\text{Ret}}} \widetilde{\mathcal{L}}_{\text{V}} = \widetilde{\mathcal{L}}_{\text{win}} + \lambda_v\left\|V_{\text{Ret}}-V_{\text{Ret}}^0\right\|_F^2~.
\end{equation}

\textbf{Closed-form solution.} This is a linear least-squares problem with Tikhonov regularization. Define $X = A_{\text{win}}^{(\text{Ret})}$, $Y = A_{\text{win}}^{(\text{full})} V_{\text{full}}$ and the $c \times c$ identity matrix $I_c$. Setting the gradient with respect to $V_{\text{Ret}}$ to zero yields the closed-form solution:
\begin{equation}
  V_{\text{Ret}}^* = \big(X^\top X + \lambda_v I_c\big)^{-1}
  \big(X^\top Y + \lambda_v V_{\text{Ret}}^0\big)~.
  \label{eq:eq7}
\end{equation}
Here, $V_{\text{Ret}}^*$ corresponds to the merged value cache.

\textbf{Dual solution via Woodbury.} When the cache budget $c$ is large, directly inverting the $c \times c$ matrix in Eq.~\ref{eq:eq7} can be computationally expensive. However, since the query window size $m$ is typically much smaller than $c$, we use the Woodbury identity to solve an $m \times m$ system instead:
\begin{equation}
V_{\text{Ret}}^* = V_{\text{Ret}}^0 + X^\top Z,
\end{equation}
where $Z = (X X^\top + \lambda_v I_m)^{-1}(Y - X V_{\text{Ret}}^0)$. In implementation, we adopt the corresponding \emph{dual} formulation to improve efficiency. Since $m$ (e.g., 32) is typically far smaller than $c$ (e.g., 10\% of the context length), this dual formulation substantially reduces both compute and memory overhead.

\textbf{Key step.}
With the current values $V_{\text{Ret}}$ fixed, we update $K_{\text{Ret}}$ by minimizing:
\begin{equation}
\min_{K_{\text{Ret}}} \widetilde{\mathcal{L}}_{\text{K}} = \widetilde{\mathcal{L}}_{\text{win}} + \lambda_k\left\|K_{\text{Ret}}-K_{\text{Ret}}^0\right\|_F^2~.
\end{equation}

\textbf{Linearized dual solution.} Unlike the value step, the key-step objective is nonlinear because $K_{\text{Ret}}$ appears inside the softmax. GRKV linearizes the attention output $f(K_{\text{Ret}})=A_{\text{win}}^{(\text{Ret})}V_{\text{Ret}}$ around the current $K_{\text{Ret}}$ and solves the resulting ridge problem in dual form. Let $E=Y-f(K_{\text{Ret}})$, $G=K_{\text{Ret}}-K_{\text{Ret}}^0$, and $J=\left.\frac{\partial f}{\partial K}\right|_{K=K_{\text{Ret}}}$, where the matrices are vectorized when applying $J$. Define $e=\operatorname{vec}(E)$ and $g=\operatorname{vec}(G)$. The key update is:
\begin{equation}
  \operatorname{vec}(K_{\text{Ret}}^*)=\operatorname{vec}(K_{\text{Ret}}^0)+J^\top\alpha,
  \label{eq:key_dual_update}
\end{equation}
where $\alpha = (JJ^\top+\lambda_k I_{\mathcal{Y}})^{-1}(e+Jg)$. Here, $I_{\mathcal{Y}}$ is the $md \times md$ identity matrix and $K_{\text{Ret}}^*$ corresponds to the merged key cache. Although the dual system has dimension $md$, it is smaller than the primal key space of dimension $cd$ when $m \ll c$.

\begin{table*}[t]
\centering
\small
\setlength{\tabcolsep}{2.4pt}
\renewcommand{\arraystretch}{1.08}

\begin{adjustbox}{max width=\textwidth}
\begin{tabular}{l*{18}{c}}
\toprule
\multirow{2}{*}{Method}
& \multicolumn{3}{c}{Single-Document QA}
& \multicolumn{3}{c}{Multi-Document QA}
& \multicolumn{3}{c}{Summarization}
& \multicolumn{3}{c}{Few-shot Learning}
& \multicolumn{2}{c}{Synthetic}
& \multicolumn{2}{c}{Code}
& \multirow{2}{*}{Avg.}
& \multirow{2}{*}{Better} \\
\cmidrule(lr){2-4}\cmidrule(lr){5-7}\cmidrule(lr){8-10}\cmidrule(lr){11-13}\cmidrule(lr){14-15}\cmidrule(lr){16-17}
& \rotatebox[origin=c]{45}{NtrvQA}
& \rotatebox[origin=c]{45}{Qasper}
& \rotatebox[origin=c]{45}{MF-en}
& \rotatebox[origin=c]{45}{Hotpot}
& \rotatebox[origin=c]{45}{2WikiQA}
& \rotatebox[origin=c]{45}{Musique}
& \rotatebox[origin=c]{45}{GovRep}
& \rotatebox[origin=c]{45}{QMSum}
& \rotatebox[origin=c]{45}{MultiNews}
& \rotatebox[origin=c]{45}{TREC}
& \rotatebox[origin=c]{45}{TriviaQA}
& \rotatebox[origin=c]{45}{SAMSum}
& \rotatebox[origin=c]{45}{PCount}
& \rotatebox[origin=c]{45}{PR-en}
& \rotatebox[origin=c]{45}{Lcc}
& \rotatebox[origin=c]{45}{RB-P} \\
\midrule

\multicolumn{19}{c}{Llama-3.1-8B-Instruct, 10\% Cache Budget} \\
\midrule
Full Cache & 30.46 & 47.29 & 55.03 & 58.65 & 51.61 & 33.03 & 35.03 & 24.97 & 27.04 & 29.00 & 84.81 & 39.29 & 10.15 & 100.00 & 51.43 & 44.19 & 45.12 & -- \\
\midrule
SnapKV     & 24.15 & 20.36 & 25.41 & 46.87 & 26.67 & 18.95 & 24.79 & 20.37 & 20.76 & 34.50 & 80.08 & 38.65 & 8.00 & 59.00 & 50.05 & 44.72 & 33.96 & 0/16 \\
w. CaM     & 23.86 & 20.15 & 26.15 & 48.79 & 26.12 & 20.20 & 23.93 & 20.35 & 20.33 & 24.50 & 79.88 & 32.16 & 4.53 & 59.50 & 43.35 & 42.48 & 32.27 & 4/16 \\
w. D2O     & 17.63 & 13.24 & 18.53 & 35.80 & 17.99 & 12.44 & 22.05 & 17.73 & 18.83 & 21.50 & 80.71 & 35.97 & 5.50 & 29.00 & 46.50 & 45.68 & 27.44 & 2/16 \\
w. AsymKV  & 19.91 & 22.35 & 27.31 & 40.18 & 25.05 & 13.66 & 24.92 & 21.13 & 23.25 & 19.00 & 87.94 & 20.59 & 2.83 & 21.00 & 48.28 & 46.12 & 28.97 & 7/16 \\
\rowcolor[RGB]{225,244,252}
w. GRKV    & 24.22 & 21.84 & 25.39 & 47.97 & 27.32 & 20.08 & 25.41 & 20.75 & 21.07 & 35.50 & 81.41 & 38.97 & 9.00 & 59.50 & 49.69 & 45.14 & \textbf{34.58} & \textbf{14/16} \\
\midrule
CriticalKV & 25.23 & 24.57 & 26.37 & 49.58 & 27.59 & 20.64 & 26.39 & 21.07 & 21.25 & 37.25 & 80.08 & 39.44 & 9.04 & 70.50 & 51.33 & 45.69 & 36.00 & 0/16 \\
w. CaM     & 23.47 & 21.59 & 28.89 & 50.96 & 28.93 & 22.10 & 24.96 & 21.14 & 21.11 & 24.50 & 79.88 & 31.72 & 6.75 & 77.50 & 42.53 & 42.33 & 34.27 & 6/16 \\
w. D2O     & 19.85 & 13.06 & 21.28 & 37.67 & 18.99 & 14.05 & 22.68 & 18.41 & 19.14 & 21.50 & 77.71 & 36.38 & 9.00 & 24.00 & 45.43 & 45.34 & 27.78 & 0/16 \\
w. AsymKV  & 20.17 & 20.77 & 27.41 & 35.27 & 28.42 & 11.94 & 24.30 & 20.97 & 23.57 & 14.00 & 83.19 & 21.63 & 1.52 & 14.00 & 51.58 & 45.84 & 27.79 & 6/16 \\
\rowcolor[RGB]{225,244,252}
w. GRKV    & 26.51 & 25.17 & 27.41 & 50.16 & 28.42 & 21.01 & 26.55 & 21.28 & 21.81 & 38.25 & 81.16 & 39.15 & 10.00 & 71.00 & 51.20 & 46.24 & \textbf{36.58} & \textbf{14/16} \\
\midrule

\multicolumn{19}{c}{Mistral-7B-Instruct-v0.3, 10\% Cache Budget} \\
\midrule
Full Cache & 28.38 & 40.72 & 51.68 & 49.00 & 36.21 & 27.81 & 34.23 & 25.41 & 26.93 & 55.75 & 84.93 & 20.94 & 5.05 & 98.00 & 46.92 & 53.86 & 42.86 & -- \\
\midrule
SnapKV     & 17.72 & 12.61 & 30.07 & 34.28 & 25.68 & 15.52 & 25.42 & 21.10 & 20.77 & 37.75 & 87.40 & 28.78 & 3.92 & 66.00 & 50.63 & 52.32 & 33.12 & 0/16 \\
w. CaM     & 17.54 & 13.58 & 29.71 & 33.57 & 20.26 & 14.81 & 24.48 & 20.88 & 20.59 & 29.00 & 85.30 & 20.38 & 4.30 & 62.00 & 53.07 & 51.40 & 31.30 & 3/16 \\
w. D2O     & 13.69 & 7.27 & 24.35 & 30.48 & 18.67 & 13.74 & 23.02 & 18.93 & 17.86 & 27.25 & 87.20 & 35.15 & 5.26 & 27.25 & 46.76 & 51.68 & 28.04 & 2/16 \\
w. AsymKV  & 20.48 & 17.13 & 27.35 & 40.24 & 23.38 & 16.02 & 25.76 & 20.50 & 22.96 & 42.25 & 81.17 & 18.12 & 4.83 & 32.25 & 44.69 & 48.78 & 30.37 & 8/16 \\
\rowcolor[RGB]{225,244,252}
w. GRKV    & 18.43 & 13.38 & 30.79 & 37.23 & 25.71 & 15.58 & 25.38 & 21.02 & 21.11 & 40.00 & 88.27 & 29.82 & 3.39 & 65.50 & 51.32 & 53.13 & \textbf{33.75} & \textbf{12/16} \\
\midrule
CriticalKV & 18.99 & 14.38 & 34.95 & 36.34 & 26.12 & 18.43 & 26.09 & 20.66 & 21.14 & 40.75 & 86.31 & 27.45 & 3.33 & 64.00 & 49.02 & 51.10 & 33.69 & 0/16 \\
w. CaM     & 16.69 & 14.50 & 30.53 & 41.02 & 26.12 & 17.10 & 25.07 & 21.10 & 21.32 & 36.00 & 84.37 & 31.04 & 4.12 & 58.00 & 52.23 & 51.32 & 33.16 & 8/16 \\
w. D2O     & 15.41 & 8.66 & 25.41 & 29.07 & 20.64 & 15.25 & 23.40 & 18.41 & 18.77 & 29.25 & 85.73 & 35.73 & 4.15 & 26.00 & 47.85 & 51.33 & 28.44 & 3/16 \\
w. AsymKV  & 20.38 & 19.65 & 28.51 & 36.83 & 23.15 & 14.12 & 25.90 & 20.88 & 23.45 & 45.25 & 80.67 & 16.46 & 2.83 & 26.00 & 44.80 & 48.98 & 29.87 & 6/16 \\
\rowcolor[RGB]{225,244,252}
w. GRKV    & 19.00 & 15.55 & 35.64 & 37.68 & 25.78 & 18.28 & 26.18 & 20.98 & 21.58 & 42.25 & 86.84 & 27.50 & 3.93 & 66.00 & 50.16 & 51.42 & \textbf{34.30} & \textbf{14/16} \\
\bottomrule
\end{tabular}
\end{adjustbox}
\caption{Detailed scores on all 16 LongBench tasks. The Avg. column reports the average score over all tasks, and the Better column reports the number of tasks on which a method outperforms its base eviction method.}
\label{tab:longbench_detailed_10}
\end{table*}

\textbf{Fixed tokens during optimization.} GRKV treats the retained cache as a global set of carriers, but some retained tokens should not be modified. In particular, we keep sink tokens, surrogate-window query tokens, and the $\beta=10\%$ retained tokens with the highest attention scores unchanged. These tokens often anchor attention or carry local query semantics, so modifying their KV cache can harm generation fidelity. In Sec.~\ref{sec:ablation}, we analyze this design choice and its effect on performance.

Fig.~\ref{fig:overview}(c) presents an overview of the GRKV pipeline, and Algorithm~\ref{alg:GRKV} provides detailed pseudocode. The full derivations for the key and value steps are provided in Appendix~\ref{appendix:D}.

\begin{table*}[t]
\centering
\small
\setlength{\tabcolsep}{3.0pt}
\renewcommand{\arraystretch}{1.08}

\begin{adjustbox}{max width=\textwidth}
\begin{tabular}{lccccccccccccccc}
\toprule
Method
& \rotatebox[origin=c]{45}{cwe}
& \rotatebox[origin=c]{45}{fwe}
& \rotatebox[origin=c]{45}{niah\_mk1}
& \rotatebox[origin=c]{45}{niah\_mk2}
& \rotatebox[origin=c]{45}{niah\_mk3}
& \rotatebox[origin=c]{45}{niah\_mq}
& \rotatebox[origin=c]{45}{niah\_mv}
& \rotatebox[origin=c]{45}{niah\_s1}
& \rotatebox[origin=c]{45}{niah\_s2}
& \rotatebox[origin=c]{45}{niah\_s3}
& \rotatebox[origin=c]{45}{qa\_1}
& \rotatebox[origin=c]{45}{qa\_2}
& \rotatebox[origin=c]{45}{vt}
& \rotatebox[origin=c]{45}{Avg.}
& \rotatebox[origin=c]{45}{Better} \\
\midrule

\multicolumn{16}{c}{Llama-3.1-8B-Instruct, 16K RULER, 10\% Cache Budget} \\
\midrule
Full Cache & 89.28 & 90.33 & 99.60 & 100.00 & 99.00 & 98.90 & 98.90 & 100.00 & 100.00 & 100.00 & 80.80 & 57.20 & 99.72 & 93.36 & -- \\
\midrule
SnapKV     & 3.24 & 59.00 & 23.20 & 4.00 & 1.40 & 17.00 & 15.10 & 78.40 & 54.20 & 2.40 & 21.00 & 28.00 & 49.72 & 27.44 & 0/13 \\
w. CaM     & 0.12 & 56.20 & 17.60 & 2.20 & 1.60 & 13.95 & 13.20 & 41.40 & 31.20 & 2.40 & 23.00 & 28.20 & 5.44 & 18.19 & 3/13 \\
w. D2O     & 0.02 & 57.80 & 16.40 & 0.40 & 0.20 & 12.70 & 11.25 & 79.80 & 47.60 & 2.40 & 14.60 & 23.80 & 36.60 & 23.35 & 1/13 \\
\rowcolor[RGB]{225,244,252}
w. GRKV    & 1.78 & 66.40 & 25.60 & 4.40 & 1.40 & 17.60 & 15.80 & 82.40 & 58.20 & 2.40 & 20.20 & 27.80 & 54.24 & \textbf{29.09} & \textbf{8/13} \\
\midrule
CriticalKV & 9.72 & 58.73 & 50.20 & 5.60 & 1.60 & 45.10 & 46.30 & 93.60 & 94.40 & 2.60 & 23.80 & 30.40 & 68.72 & 40.83 & 0/13 \\
w. CaM     & 0.10 & 57.87 & 34.20 & 2.20 & 1.20 & 28.65 & 28.05 & 50.60 & 71.00 & 2.60 & 25.60 & 25.80 & 6.48 & 25.72 & 1/13 \\
w. D2O     & 0.28 & 53.80 & 36.40 & 1.00 & 0.00 & 36.95 & 32.45 & 93.80 & 80.20 & 2.40 & 15.00 & 26.20 & 52.28 & 33.14 & 1/13 \\
\rowcolor[RGB]{225,244,252}
w. GRKV    & 8.80 & 64.93 & 50.00 & 5.80 & 1.60 & 45.60 & 46.65 & 93.60 & 95.40 & 2.60 & 24.00 & 31.00 & 69.60 & \textbf{41.51} & \textbf{8/13} \\
\midrule

\multicolumn{16}{c}{Mistral-7B-Instruct-v0.3, 16K RULER, 10\% Cache Budget} \\
\midrule
Full Cache & 82.64 & 88.07 & 97.60 & 95.00 & 77.60 & 88.50 & 88.65 & 94.40 & 96.40 & 99.60 & 72.00 & 49.80 & 96.04 & 86.64 & -- \\
\midrule
SnapKV     & 45.98 & 82.80 & 10.40 & 1.80 & 0.40 & 9.60 & 9.50 & 40.80 & 3.40 & 2.40 & 24.20 & 31.80 & 12.32 & 21.18 & 0/13 \\
w. CaM     & 10.32 & 64.73 & 10.20 & 1.40 & 0.00 & 9.05 & 9.20 & 5.40 & 4.00 & 2.40 & 25.40 & 27.40 & 1.44 & 13.15 & 2/13 \\
w. D2O     & 22.78 & 73.13 & 8.20 & 0.00 & 0.00 & 8.70 & 9.20 & 43.20 & 4.00 & 2.40 & 21.60 & 28.60 & 2.68 & 17.27 & 2/13 \\
\rowcolor[RGB]{225,244,252}
w. GRKV    & 47.32 & 83.33 & 10.80 & 2.00 & 0.40 & 9.80 & 9.70 & 42.20 & 4.40 & 2.40 & 24.60 & 30.40 & 14.32 & \textbf{21.67} & \textbf{10/13} \\
\midrule
CriticalKV & 59.10 & 78.20 & 13.00 & 2.40 & 0.00 & 10.05 & 9.45 & 36.80 & 9.20 & 2.40 & 28.20 & 31.40 & 13.24 & 22.57 & 0/13 \\
w. CaM     & 2.06 & 19.80 & 11.80 & 0.20 & 0.00 & 9.15 & 9.20 & 0.00 & 5.60 & 2.40 & 26.00 & 26.80 & 0.00 & 8.69 & 0/13 \\
w. D2O     & 30.74 & 71.33 & 9.80 & 0.20 & 0.00 & 8.70 & 9.50 & 40.40 & 5.80 & 2.40 & 22.00 & 29.00 & 4.56 & 18.03 & 2/13 \\
\rowcolor[RGB]{225,244,252}
w. GRKV    & 59.84 & 79.33 & 13.60 & 2.80 & 0.00 & 10.15 & 9.75 & 39.20 & 9.60 & 2.40 & 28.00 & 31.60 & 14.00 & \textbf{23.10} & \textbf{10/13} \\
\bottomrule
\end{tabular}
\end{adjustbox}
\caption{Detailed scores on all 13 RULER tasks. The Avg. column reports the average score over all tasks, and the Better column reports the number of tasks on which a method outperforms its base eviction method.}
\label{tab:ruler_detailed_10}
\end{table*}

\begin{table}[t]
\centering
\small
\setlength{\tabcolsep}{5pt}
\renewcommand{\arraystretch}{1.05}
\newcommand{\groupsep}{\addlinespace[1pt]\specialrule{0.3pt}{1pt}{1pt}}
\begin{tabular*}{\columnwidth}{@{\extracolsep{\fill}}llc@{}}
\toprule
Factor & Setting & Avg. \\
\midrule
Regularization strength
& $\lambda=1$ & 34.33 \\
($\lambda=\lambda_k=\lambda_v$)
& $\lambda=10^{-1}$ & 34.29 \\
& $\lambda=10^{-2}\dagger$ & \textbf{34.58} \\
& $\lambda=0$ & 32.06 \\
\groupsep

Fixed retained-token ratio
& $\beta=0\%$ & 34.34 \\
($\beta$)
& $\beta=10\%\dagger$ & \textbf{34.58} \\
& $\beta=20\%$ & 34.28 \\
& $\beta=30\%$ & 34.25 \\
\groupsep

Update steps
& $S=1\dagger$ & \textbf{34.58} \\
($S$)
& $S=2$ & 34.33 \\
& $S=3$ & 34.19 \\
\groupsep

Surrogate window size
& $m=32\dagger$ & \textbf{34.58} \\
($m$)
& $m=48$ & 33.86 \\
& $m=64$ & 33.49 \\
\groupsep

Sink and window tokens
& Fixed$\dagger$ & \textbf{34.58} \\
& Updated & 34.19 \\
\groupsep

Optimized cache components
& GRK & 34.21 \\
& GRV & 34.40 \\
& GRKV$\dagger$ & \textbf{34.58} \\
\bottomrule
\end{tabular*}
\caption{GRKV ablations on LongBench with SnapKV on \textit{Llama-3.1-8B-Instruct} at a 10\% cache budget. Bold indicates the best setting in each group. $\dagger$ denotes the default setting. $\lambda=0$ denotes no regularization.}
\label{tab:ablation_summary}
\end{table}

\section{Experiments}

\subsection{Experimental Settings}

\textbf{Models and Benchmarks.} We evaluate GRKV on three open-source LLMs: \textit{Mistral-7B-Instruct-v0.3}~\citep{jiang2023mistral7b}, supporting up to 32K tokens; \textit{Llama-3.1-8B-Instruct}~\citep{grattafiori2024llama} and \textit{Qwen3-14B}~\citep{yang2025qwen3}, both supporting up to 128K tokens. The evaluation uses two long-context benchmarks: LongBench~\citep{bai-etal-2024-longbench} and RULER~\citep{hsieh2024ruler}. Detailed benchmark information is provided in Appendix~\ref{appendix:E}.

\textbf{Baselines.} For span-based eviction methods, we consider SnapKV~\citep{li2024snapkv}, PyramidKV~\citep{cai2024pyramidkv}, CriticalKV~\citep{feng2025identify}, and Ada-KV~\citep{feng2024ada}. For token-based eviction methods, we consider H2O~\citep{zhang2023h2o}. For KV-cache merging methods, we evaluate CaM~\citep{zhang2024cam}, D2O~\citep{wan2024d2o}, and AsymKV~\citep{cui2025homogeneous}.

\textbf{Protocol and hyperparameters.} We follow the protocols of~\citet{feng2025identify, feng2024ada} and~\citet{devoto2025expectedattention}: for tasks with a prompt and a final query, we compress the prompt independently of the final query, which yields a more challenging and realistic setting. Consistent with~\citet{zhang2024cam}, each merging method is paired with a base eviction method to test whether merging can recover discarded information. For GRKV, we use one alternating KV update step ($S=1$), set the surrogate window size to $m=32$, as in prior KV-cache eviction methods~\citep{li2024snapkv, feng2025identify, feng2024ada}, and keep sink tokens, surrogate-window query tokens, and the top $\beta=10\%$ retained tokens by attention score fixed. We set $\lambda_k=\lambda_v=10^{-2}$ for Llama and $\lambda_k=\lambda_v=10^{-1}$ for Mistral and Qwen. We use FlashAttention-2~\citep{dao2022flashattention, dao2023flashattention} for all methods except AsymKV, which cannot leverage FlashAttention-2.

\subsection{Experimental Results}
\textbf{LongBench.} LongBench~\citep{bai-etal-2024-longbench} contains 16 long-context tasks spanning six categories: single-/multi-document QA, summarization, few-shot learning, synthetic tasks, and code. We evaluate under a 10\% cache budget and report per-task scores in Table~\ref{tab:longbench_detailed_10}. GRKV consistently improves both base eviction methods across both model backbones. On \textit{Llama-3.1-8B-Instruct}, it raises the average score from 33.96 to 34.58 with SnapKV and from 36.00 to 36.58 with CriticalKV, improving 14/16 tasks in both cases. On \textit{Mistral-7B-Instruct-v0.3}, it improves SnapKV from 33.12 to 33.75 and CriticalKV from 33.69 to 34.30, with gains on 12/16 and 14/16 tasks. Other merging baselines are less reliable. CaM yields occasional gains but often reduces the average score, while D2O and AsymKV frequently degrade under span-based retention. In contrast, GRKV consistently improves the average score in all settings, demonstrating the benefit of global, ridge-regularized reconstruction. Across three seeds, GRKV yields positive gains over the models and eviction methods. Complete statistics are reported in Appendix~\ref{sec:multiseed_results}.

\textbf{RULER.} RULER~\citep{hsieh2024ruler} evaluates long-context understanding across extraction (CWE/FWE), retrieval (NIAH), question answering, and variable tracking (VT) tasks. Under a 10\% cache budget, Table~\ref{tab:ruler_detailed_10} shows that GRKV consistently improves both base eviction methods across both model backbones. Because AsymKV incurs high overhead at 16K and increases latency by $\sim$15×, we exclude it. On \textit{Llama-3.1-8B-Instruct}, GRKV raises the average score from 27.44 to 29.09 with SnapKV and from 40.83 to 41.51 with CriticalKV, improving 8/13 tasks in both cases. On \textit{Mistral-7B-Instruct-v0.3}, it improves SnapKV from 21.18 to 21.67 and CriticalKV from 22.57 to 23.10, with gains on 10/13 tasks in both cases. In contrast, CaM and D2O often degrade performance, especially on retrieval tasks, due to over-merging.

Additional results on LongBench and RULER under a 20\% cache budget are given in Appendix~\ref{appendix:F}.

\subsection{Ablations, Compatibility, and Efficiency}
\label{sec:ablation}

All ablations use SnapKV with GRKV under a 10\% cache budget and report LongBench averages.

\textbf{Regularization strength.} As shown in Table~\ref{tab:ablation_summary}, a mild ridge penalty is important: $\lambda_k=\lambda_v=10^{-2}$ performs best (34.58), while removing regularization drops the score to 32.06. This indicates that regularization prevents overly aggressive KV updates while still allowing useful reconstruction.

\textbf{Fixed retained-token ratio.} Table~\ref{tab:ablation_summary} shows that fixing a small fraction of high-attention retained tokens is beneficial. The default $\beta=10\%$ performs best (34.58), while fixing no retained tokens is weaker (34.34) and larger ratios slightly reduce performance by constraining too many carriers.

\textbf{Update steps.} Table~\ref{tab:ablation_summary} shows that one alternating KV update step is sufficient. Increasing $S$ from 1 to 2 or 3 lowers the average score, suggesting that repeated updates can overfit the surrogate window.

\textbf{Surrogate window size.} Table~\ref{tab:ablation_summary} shows that the default $m=32$ achieves the best score in this sweep (34.58), while increasing the window size to $m=48$ and $m=64$ lowers the average to 33.86 and 33.49, respectively. This supports using $m=32$, which is also consistent with prior eviction baselines~\citep{li2024snapkv, feng2025identify, feng2024ada}.

\textbf{Fixed sink and window tokens.} Table~\ref{tab:ablation_summary} validates keeping sink and surrogate-window query tokens fixed: allowing them to be updated reduces performance from 34.58 to 34.19. Preserving these special tokens improves optimization stability.

\textbf{Optimized cache components.} Table~\ref{tab:ablation_summary} compares updating only keys (GRK), only values (GRV), and both (GRKV). Updating both components performs best (34.58), indicating that key and value corrections provide complementary benefits. The key update accounts for 84.6--84.7\% of added regression time; GRV is therefore a lightweight option when prefill latency is critical (Appendix~\ref{sec:controlled_efficiency}).

\textbf{Loss--performance relationship.} Reconstruction discrepancy is negatively correlated with LongBench/RULER scores, but not strictly monotonically; Appendix~\ref{sec:loss_performance} gives the full analysis.

\textbf{Compatibility.} Table~\ref{tab:compatibility_longbench} shows that GRKV improves diverse eviction backbones under a 10\% cache budget on LongBench. On \textit{Llama-3.1-8B-Instruct}, it improves PyramidKV, which dynamically allocates KV-cache budgets across layers, from 34.40 to 35.21, and Ada-KV, which dynamically allocates KV-cache budgets across heads, from 34.72 to 35.28. It also benefits the token-based retention baseline H2O, improving it from 29.65 to 31.68. On the larger \textit{Qwen3-14B} model, GRKV improves span-based SnapKV from 38.29 to 38.98 and CriticalKV from 41.68 to 42.20.

\begin{figure*}[t]
\centering
\includegraphics[width=\textwidth]{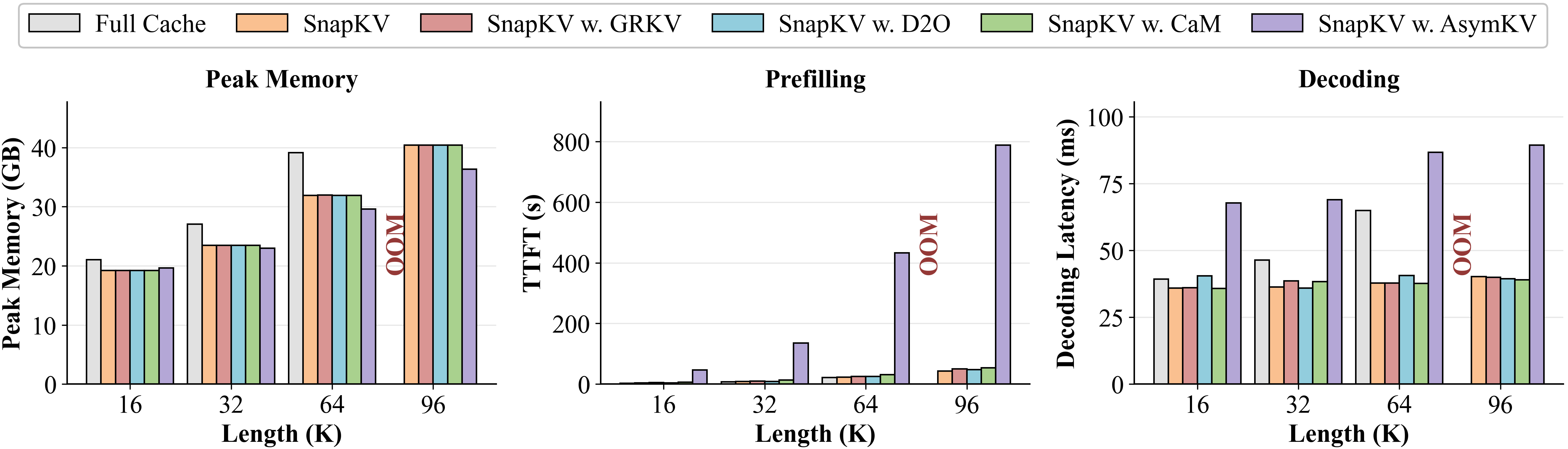}
\caption{Peak memory, TTFT (time to first token), and decoding latency measured across 16K--96K context lengths.}
\label{fig:kv_efficiency}
\end{figure*}

\textbf{Efficiency.} We profile \textit{Llama-3.1-8B-Instruct} on an A6000 with 16K--96K contexts at a 10\% cache budget. As shown in Fig.~\ref{fig:kv_efficiency}, compression methods substantially reduce peak memory and avoid the full-cache out-of-memory failure at 96K. SnapKV with GRKV closely follows SnapKV in memory usage and retains efficient decoding: at 64K, it reduces latency from 64.84 to 37.75~ms/token compared with Full Cache, while remaining close to SnapKV. Its prefill overhead is moderate compared with heavier merging methods; at 96K, SnapKV with GRKV takes 49.71~s, close to D2O (48.20~s) and below CaM (53.87~s), whereas AsymKV reaches 788.49~s and also has higher decoding latency because it cannot use FlashAttention-2. The budget denotes \emph{persistent} KV storage; GRKV transiently accesses only the current layer's full cache. Controlled tests find the same peak memory as SnapKV and 0.8--2.6\% additional TTFT under unified eager attention. Appendix~\ref{sec:controlled_efficiency} reports the complete memory, latency, and iso-resource trade-offs.

Additional RULER ablations, compatibility results, and efficiency tests across GPUs, batch sizes, and context lengths are provided in Appendix~\ref{appendix:G}.

\begin{table}[t]
\centering
\small
\setlength{\tabcolsep}{5pt}
\renewcommand{\arraystretch}{1.08}
\begin{tabular*}{\columnwidth}{@{\extracolsep{\fill}}lllcc@{}}
\toprule
Model & Retention & Method & Base & GRKV \\
\midrule
\multirow{3}{*}{\shortstack[l]{\textit{Llama-3.1}\\\textit{-8B-Instruct}}}
& \multirow{2}{*}{Span-based}
& PyramidKV & 34.40 & \textbf{35.21} \\
& & Ada-KV & 34.72 & \textbf{35.28} \\
\addlinespace[1pt]
\cmidrule{2-5}
\addlinespace[1pt]
& Token-based
& H2O & 29.65 & \textbf{31.68} \\
\midrule
\multirow{2}{*}{\textit{Qwen3-14B}}
& \multirow{2}{*}{Span-based}
& SnapKV & 38.29 & \textbf{38.98} \\
& & CriticalKV & 41.68 & \textbf{42.20} \\
\bottomrule
\end{tabular*}
\caption{LongBench compatibility at a 10\% cache budget. Each row compares a base eviction method with its GRKV-enhanced variant.}
\label{tab:compatibility_longbench}
\end{table}

\section{Conclusion}

We proposed GRKV, a general training-free KV-cache merging method for modern span-based KV-cache eviction methods. GRKV formulates merging as a global regression objective to minimize the attention-output discrepancy between a compressed cache and the full cache. By treating all retained tokens as carriers and solving a ridge-regression problem, GRKV more effectively recovers information from evicted tokens and mitigates over-merging caused by local heuristics. Across 16 LongBench and 13 RULER tasks, GRKV is the only KV-cache merging method that improves overall performance with minimal overhead, suggesting that aligning compression objectives with model outputs can yield practical inference gains.

\section*{Limitations}

Our empirical evaluation covers three open-source models: \textit{Llama-3.1-8B-Instruct}, \textit{Mistral-7B-Instruct-v0.3}, and \textit{Qwen3-14B}, and evaluates them on LongBench and RULER. These benchmarks mainly evaluate English long-context understanding, code-oriented tasks, and synthetic retrieval tasks. We therefore do not claim that the same gains will necessarily generalize to other model families, larger proprietary systems, or multilingual/multimodal settings.

The prompt-derived surrogate window is an empirical proxy for future queries, not a guarantee across all heads, samples, tasks, or window positions. Weak surrogate--actual-query alignment can cause the reconstruction update not to transfer, and the task-level gains are not uniform. When prefill latency is the primary constraint, increasing the cache budget may be preferable.

\section*{Acknowledgments}

We would like to thank the anonymous reviewers for their valuable feedback and constructive comments. We sincerely thank Mingwei Xu and Wanyun Cui for their valuable technical guidance during our re-implementation of AsymKV~\citep{cui2025homogeneous}. This work was supported by the Major Key Project of PCL (PCL2024A06), the National Key Research and Development Program of China (2024YFA1011900), the HPC platform of ShanghaiTech University, the Core Facility Platform of Computer Science and Communication, SIST, ShanghaiTech University, and the Key Areas Research plan of the Guangdong S\&T programme (2025B0101120008).

\bibliography{custom,anthology-1,anthology-2}

\appendix

\section{Why Span-Based Retention Amplifies Over-Merging}
\label{appendix:A}

This appendix provides a first-order analysis of why span-based retention amplifies over-merging in local KV-cache merging methods. We show that span-based retention does not merely change which tokens are retained; under local matching rules, it reduces the number of active carrier tokens. This simultaneously increases the load on each carrier and shrinks the first-order feasible space for reconstructing the full-cache attention output.

\subsection{Setup and Notation}

Consider a single attention head. Let the full cache be denoted by $K_{\text{full}},V_{\text{full}}\in\mathbb{R}^{n\times d}$. After eviction, the retained cache is $K_{\text{Ret}}^0,V_{\text{Ret}}^0\in\mathbb{R}^{c\times d}$, and the evicted cache is represented by $K_{\text{evict}},V_{\text{evict}}\in\mathbb{R}^{e\times d}$, where $e=n-c$. For the surrogate query window
$Q_{\text{win}}\in\mathbb{R}^{m\times d}$, define the full-cache target $Y = A_{\text{win}}^{(\text{full})}V_{\text{full}}$, where $A_{\text{win}}^{(\text{full})} = \operatorname{softmax}(Q_{\text{win}}K_{\text{full}}^\top / \sqrt{d})$. For any retained cache $(K,V)$, define the attention output $F(K,V) = A(Q_{\text{win}},K)V$, where $A(Q_{\text{win}},K) = \operatorname{softmax}(Q_{\text{win}}K^\top / \sqrt{d})$. The window-level objective in the main text is:
\begin{equation}
  \widetilde{\mathcal{L}}_{\text{win}}(K,V)
  =
  \left\|Y-F(K,V)\right\|_F^2 .
\end{equation}

\subsection{Local KV Merging and Carrier Support}
Many local KV-cache merging methods can be abstracted to first order as carrier-restricted updates of the retained KV cache:
\begin{equation}
\begin{aligned}
K_{\text{Ret}}^* &= K_{\text{Ret}}^0 + B_KK_{\text{evict}}, \\
V_{\text{Ret}}^* &= V_{\text{Ret}}^0 + B_VV_{\text{evict}},
\end{aligned}
\label{eq:appendixA_local_kv_merge}
\end{equation}
where $B_K,B_V\in\mathbb{R}^{c\times e}$ are merge-assignment matrices and $K_{\text{Ret}}^*$, $V_{\text{Ret}}^*$ correspond to the merged key cache and merged value cache, respectively. Local adjacency-based or key-similarity-based matching rules usually assign each evicted token to one or a small number of candidate carriers. In the one-carrier case, if $\pi(j)$ is the retained carrier selected for the evicted token $j$, then:
\begin{equation}
(B_K)_{ij}=(B_V)_{ij}=0
\quad\text{whenever}\quad
i\neq \pi(j).
\label{eq:appendixA_one_carrier}
\end{equation}
More generally, local KV-cache matching rules restrict the nonzero rows of $B_K$ and $B_V$ to an active carrier set $\mathcal{C}\subseteq\{1,\ldots,c\}$.

\paragraph{Claim 1: carrier-load amplification.}
Let $b=|\mathcal{C}|$, and let $\ell_i = \left|\{j:\pi(j)=i\}\right|$ for $i\in\mathcal{C}$ denote the number of evicted tokens assigned to carrier $i$. Since $\sum_{i\in\mathcal{C}}\ell_i=e$, the pigeonhole principle gives: 
\begin{equation}
\max_{i\in\mathcal{C}}\ell_i
\ge
\left\lceil\frac{e}{b}\right\rceil~.
\label{eq:appendixA_load_bound}
\end{equation}
With token-based retention, retained tokens are typically dispersed, whereas span-based retention forms contiguous spans and matches evicted tokens in the gaps mainly to span boundaries. Its empirical premise is that the tested span-based methods use fewer active carriers than token-based retention at the same merge scale, $b_{\mathrm{span}}<b_{\mathrm{token}}$ (Appendix~\ref{sec:carrier_quantitative}). Equation~\eqref{eq:appendixA_load_bound} then gives a larger maximum-load lower bound, $\lceil e/b_{\mathrm{span}}\rceil>\lceil e/b_{\mathrm{token}}\rceil$, under these tested local matching rules.

This load amplification helps explain over-merging. For a boundary carrier $i$, Eq.~\eqref{eq:appendixA_local_kv_merge} gives:
\begin{equation}
  \begin{aligned}
  \Delta k_i &= \sum_{j=1}^{e}(B_K)_{ij}k_j^{(\text{evict})}, \\
  \Delta v_i &= \sum_{j=1}^{e}(B_V)_{ij}v_j^{(\text{evict})}.
  \end{aligned}
  \label{eq:appendixA_row_updates}
\end{equation}
As more evicted tokens are assigned to the same carrier, both $\Delta k_i$ and $\Delta v_i$ accumulate more terms. Unless these terms cancel, the carrier can move farther from its post-eviction representation. Large value updates blur the content stored in the carrier, while large key updates perturb the attention logits:
\begin{equation}
  \Delta s_{ri} = \frac{q_r^\top \Delta k_i}{\sqrt{d}},
\end{equation}
thereby changing how query $q_r$ attends to that carrier. Thus, span-based retention amplifies over-merging in both key and value spaces.

\subsection{First-Order Bottleneck for KV Reconstruction}
The load argument explains why boundary carriers become prone to over-merging. We next show that this concentration also restricts the optimization geometry of the objective used in the main text.

\paragraph{Claim 2: first-order carrier bottleneck.}
Let $A_0 = A(Q_{\text{win}},K_{\text{Ret}}^0)$, $Y_0 = F(K_{\text{Ret}}^0,V_{\text{Ret}}^0) = A_0V_{\text{Ret}}^0$, and $R=Y-Y_0$ be the residual that merging should reconstruct. For small updates $\Delta K$ and $\Delta V$, the first-order expansion of $F$ around the post-eviction cache is:
\begin{equation}
\begin{aligned}
F(K_{\text{Ret}}^0+\Delta K,V_{\text{Ret}}^0+\Delta V) = &Y_0 + \mathcal{J}_K[\Delta K] \\
&+ A_0\Delta V + \mathcal{R},
\end{aligned}
\label{eq:appendixA_linearization}
\end{equation}
where $\mathcal{J}_K$ is the first-order derivative operator of $A(Q_{\text{win}},K)V_{\text{Ret}}^0$ with respect to $K$, evaluated at $K_{\text{Ret}}^0$, and $\|\mathcal{R}\|_F=\mathcal{O}(\|\Delta K\|_F\|\Delta V\|_F+\|\Delta K\|_F^2)$. This is the same type of local linearization used by the key step in the main text; at later alternating steps, the same argument applies with the current value cache replacing $V_{\text{Ret}}^0$. Let $J_K\in\mathbb{R}^{md\times cd}$ denote the vectorized matrix of $\mathcal{J}_K$.

Let $S_{\mathcal{C}}\in\mathbb{R}^{c\times b}$ select the active carrier rows. A local merge that uses only carriers in $\mathcal{C}$ satisfies:                                               
\begin{equation}
  \Delta K=S_{\mathcal{C}}U_K,
  \qquad
  \Delta V=S_{\mathcal{C}}U_V,
\end{equation}
where $U_K,U_V\in\mathbb{R}^{b\times d}$. After vectorization, the first-order change in the pre-projection attention output lies in the subspace: 
\begin{equation}
  \mathcal{T}_{\mathcal{C}}
  =
  \operatorname{col}\!\left(
  \left[
  J_K(I_d\otimes S_{\mathcal{C}})
  \quad
  I_d\otimes(A_0S_{\mathcal{C}})
  \right]\right)
  \subseteq\mathbb{R}^{md}.
  \label{eq:appendixA_tangent_space}
\end{equation}
Therefore:
\begin{equation}
  \dim(\mathcal{T}_{\mathcal{C}})
  \le
  \min(md,2bd) .
  \label{eq:appendixA_dim_bound}
\end{equation}
The best unregularized first-order reconstruction attainable by any local KV merge using only $\mathcal{C}$ is the projection of $\operatorname{vec}(R)$ onto this subspace:
\begin{equation}
  \begin{aligned}
  \min_{U_K,U_V}
  &\left\|
  \operatorname{vec}(R)
  -
  \operatorname{vec}\!\left(\mathcal{J}_K[S_{\mathcal{C}}U_K]+A_0S_{\mathcal{C}}U_V\right)
  \right\|_2^2 \\
  =
  &\left\|
  (I-\Pi_{\mathcal{T}_{\mathcal{C}}})\operatorname{vec}(R)
  \right\|_2^2 ,
  \end{aligned}
  \label{eq:appendixA_projection_error}
\end{equation}
where $\Pi_{\mathcal{T}_{\mathcal{C}}}$ is the orthogonal projector onto $\mathcal{T}_{\mathcal{C}}$. Eq.~\eqref{eq:appendixA_projection_error} formalizes the bottleneck: any residual component outside $\mathcal{T}_{\mathcal{C}}$ cannot be recovered by local KV updates restricted to the active carriers.

If all retained tokens are allowed to act as carrier tokens, the corresponding subspace is $\mathcal{T}_{\text{all}}$ with dimension at most: 
\begin{equation}
\dim(\mathcal{T}_{\text{all}})
  \le
  \min(md,2cd) .
\end{equation}
Since $\mathcal{C}\subseteq\{1,\ldots,c\}$ and $\mathcal{T}_{\mathcal{C}}\subseteq\mathcal{T}_{\text{all}}$, we have:
\begin{equation}
  \left\|
  (I-\Pi_{\mathcal{T}_{\mathcal{C}}})\operatorname{vec}(R)
  \right\|_2^2
  \ge
  \left\|
  (I-\Pi_{\mathcal{T}_{\text{all}}})\operatorname{vec}(R)
  \right\|_2^2~.
  \label{eq:appendixA_monotone_error}
\end{equation}
Thus, even when both keys and values are updated, restricting updates to an active carrier subset is at most as expressive as updating all retained tokens, reducing the first-order degrees of freedom from the all-carrier upper bound $2cd$ to $2bd$. The measured condition $b_{\mathrm{span}}<b_{\mathrm{token}}$ makes this bottleneck stronger for the tested span-based methods than for token-based retention. Carrier concentration cannot improve the best local first-order reconstruction and is strictly worse when the residual has components captured by $\mathcal{T}_{\text{all}}$ but not by $\mathcal{T}_{\mathcal{C}}$.

\subsection{Connection to GRKV}

GRKV is designed to avoid the two failure modes above. First, it treats the retained cache as a global carrier set rather than preselecting only boundary carriers. In the notation above, GRKV uses the all-carrier space $\mathcal{T}_{\text{all}}$ rather than the boundary-restricted space $\mathcal{T}_{\mathcal{C}}$, which expands the feasible first-order reconstruction space for both $K_{\text{Ret}}$ and $V_{\text{Ret}}$. Second, GRKV directly optimizes the window-level attention-output discrepancy:
\begin{equation}
  \widetilde{\mathcal{L}}_{\text{win}}
  =
  \left\|                                    
  A_{\text{win}}^{(\text{full})}V_{\text{full}}
  -
  A_{\text{win}}^{(\text{Ret})}V_{\text{Ret}}
  \right\|_F^2,
\end{equation}
instead of relying on local token matching. The value step solves a global ridge-regression problem, and the key step applies a locally linearized ridge update analyzed above.
\begin{equation}
\begin{aligned}
  &\lambda_k\|K_{\text{Ret}}-K_{\text{Ret}}^0\|_F^2
  +
  \lambda_v\|V_{\text{Ret}}-V_{\text{Ret}}^0\|_F^2 \\
  &=
  \sum_{i=1}^{c}
  \lambda_k\|\Delta k_i\|_2^2
  +
  \lambda_v\|\Delta v_i\|_2^2~.
\end{aligned}
\end{equation}
The regularizers penalize excessive movement of any carrier in both key and value spaces. Consequently, GRKV enlarges the carrier set and controls update magnitude, mitigating the over-merging that span-based retention induces under local merging.

\begin{table*}[t]
\centering
\scriptsize
\setlength{\tabcolsep}{3.1pt}
\renewcommand{\arraystretch}{1.06}
\begin{adjustbox}{width=\textwidth}
\begin{tabular}{llcccccccc}
\toprule
Model / Dataset & Budget
& \multicolumn{2}{c}{$b/c$}
& \multicolumn{2}{c}{$b_{\mathrm{eff}}/c$}
& \multicolumn{2}{c}{Gini}
& \multicolumn{2}{c}{$A_{\max}$} \\
\cmidrule(lr){3-4}\cmidrule(lr){5-6}\cmidrule(lr){7-8}\cmidrule(lr){9-10}
& & H2O & Span & H2O & Span & H2O & Span & H2O & Span \\
\midrule
Llama / NarrativeQA & 10\% & 0.84 & 0.68 / 0.70 & 0.33 & 0.19 / 0.20 & 0.62 & 0.75 / 0.74 & 14.42$\times$ & 33.92$\times$ / 32.25$\times$ \\
Llama / NarrativeQA & 20\% & 0.69 & 0.55 / 0.58 & 0.29 & 0.16 / 0.17 & 0.67 & 0.78 / 0.77 & 18.59$\times$ & 47.73$\times$ / 42.23$\times$ \\
Mistral / HotpotQA & 10\% & 0.89 & 0.80 / 0.80 & 0.41 & 0.22 / 0.25 & 0.56 & 0.69 / 0.67 & 10.37$\times$ & 32.64$\times$ / 29.30$\times$ \\
Mistral / HotpotQA & 20\% & 0.75 & 0.57 / 0.60 & 0.36 & 0.18 / 0.20 & 0.62 & 0.76 / 0.75 & 13.17$\times$ & 33.39$\times$ / 32.20$\times$ \\
\bottomrule
\end{tabular}
\end{adjustbox}
\caption{Active-carrier support and assignment concentration. ``Span'' reports SnapKV/CriticalKV under the same local matching rule.}
\label{tab:carrier_concentration}
\end{table*}

\begin{table*}[t]
\centering
\scriptsize
\setlength{\tabcolsep}{4.0pt}
\renewcommand{\arraystretch}{1.06}
\begin{tabular*}{\textwidth}{@{\extracolsep{\fill}}llcccc@{}}
\toprule
Model / Dataset & Budget & Boundary carrier fraction & Boundary assignment ratio & Enrichment & Boundary/interior load \\
\midrule
Llama / NarrativeQA & 10\% & 20\% / 21\% & 49\% / 49\% & 2.45$\times$ / 2.33$\times$ & 4.26$\times$ / 3.97$\times$ \\
Llama / NarrativeQA & 20\% & 17\% / 18\% & 46\% / 46\% & 2.71$\times$ / 2.56$\times$ & 4.53$\times$ / 4.40$\times$ \\
Mistral / HotpotQA & 10\% & 18\% / 19\% & 40\% / 39\% & 2.22$\times$ / 2.05$\times$ & 3.17$\times$ / 3.09$\times$ \\
Mistral / HotpotQA & 20\% & 16\% / 16\% & 44\% / 43\% & 2.75$\times$ / 2.69$\times$ & 4.42$\times$ / 4.37$\times$ \\
\bottomrule
\end{tabular*}
\caption{Span-boundary overload. Slash-separated values denote SnapKV/CriticalKV.}
\label{tab:boundary_overload}
\end{table*}

\begin{figure*}[!p]
  \centering
  \includegraphics[
    width=\linewidth,
    height=\textheight,
    keepaspectratio
  ]{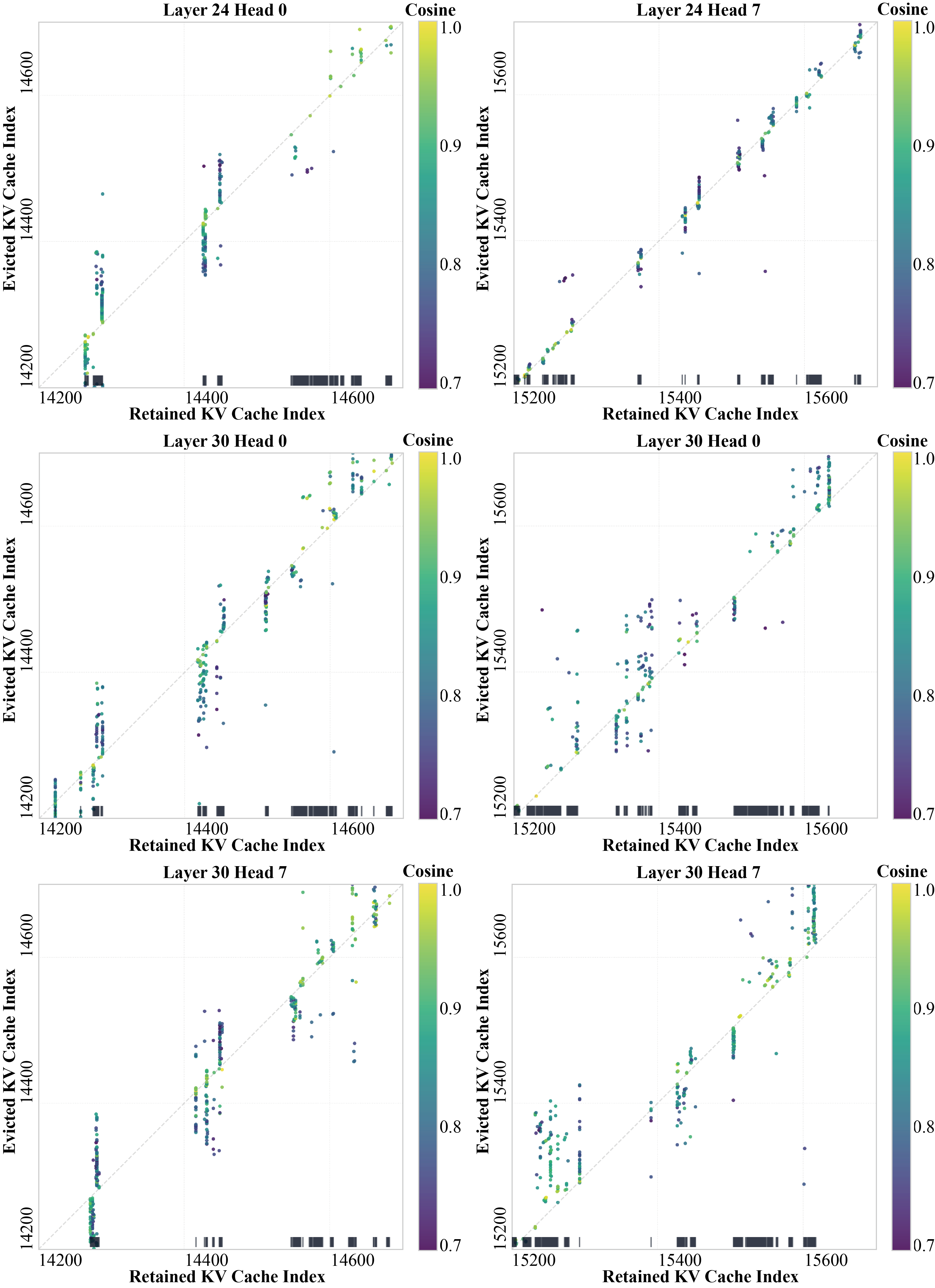}
  \caption{\textbf{Additional KV merge maps on \textsc{HotpotQA}.}
  We visualize key-similarity merge assignments (evicted $\rightarrow$ retained) after SnapKV's span-based retention across multiple layers and heads. Each point denotes a matched token pair, and the color indicates the cosine similarity between their key vectors. The black bars on the \textit{x-axis} mark the retained spans. The left and right columns correspond to two representative context regions from the same sample. They show that merge assignments remain concentrated around span boundaries across heads and layers.}
  \label{fig:kv_merge_map_hotpotqa}
\end{figure*}

\section{Additional KV Merge Map Visualizations}
\label{appendix:B}

To complement the \textsc{NarrativeQA} example in \cref{fig:kv_merge_map}, we provide additional KV merge-map visualizations on \textsc{HotpotQA}~\citep{yang-etal-2018-hotpotqa}. We follow the same setup as in the main text: SnapKV performs span-based retention to determine the retained KV cache, and a representative \emph{key-similarity} matching strategy (as in D2O/KVMerger) assigns each evicted token to the retained token whose key has the highest cosine similarity. The resulting correspondence induces a merge in which evicted values are fused into their matched retained carriers.

\paragraph{How to read the merge map.}
In each subplot of \cref{fig:kv_merge_map_hotpotqa}, the \textit{x-axis} gives the original token index of the retained token, and the \textit{y-axis} gives the original token index of the evicted token merged into the retained cache. Each point represents one matched pair $(\text{evicted} \rightarrow \text{retained})$, with color encoding the cosine similarity between their keys. The dashed diagonal is a visual reference: assignments near the diagonal correspond to merges onto nearby retained tokens in the original sequence. Finally, the black bars on the \textit{x-axis} indicate the spans retained by SnapKV, highlighting the span structure imposed by modern eviction methods.

\paragraph{Concentration induced by span-based retention.}
Across layers (e.g., 24 vs.\ 30), heads (e.g., 0 vs.\ 7), and two representative context regions corresponding to the left and right columns, \cref{fig:kv_merge_map_hotpotqa} exhibits the same qualitative behavior as \cref{fig:kv_merge_map}: once retention becomes span-based, merge assignments become strongly \emph{concentrated}. Rather than being distributed across many carriers, a large number of evicted tokens are funneled into a small subset of retained tokens, appearing as prominent vertical bands (many y-values sharing the same x-value). These bands align closely with span boundaries (i.e., near the first and last retained tokens within each black-bar segment), whereas interior tokens within retained spans receive substantially fewer assignments. This visualization supports our main-text claim that span-based retention reshapes merge assignments into an imbalanced carrier load, in which a few boundary carrier tokens are forced to absorb a disproportionate fraction of evicted information---thereby increasing the risk of over-merging and representation blurring under local matching rules.

\subsection{Quantitative Carrier Concentration}
\label{sec:carrier_quantitative}

We compare token-based H2O with span-based SnapKV and CriticalKV on \textit{Llama-3.1-8B-Instruct}/\textsc{NarrativeQA} and \textit{Mistral-7B-Instruct-v0.3}/\textsc{HotpotQA} at 10\% and 20\% cache budgets. All methods use the same local candidate range and key-cosine matching rule. Within each sample--layer--KV-head unit, let $a_i$ be the number of evicted tokens assigned to retained token $i$, and define
\begin{equation}
\begin{aligned}
b&=\sum_{i=1}^{c}\mathbf{1}[a_i>0],\\
b_{\mathrm{eff}}&=\frac{(\sum_i a_i)^2}{\sum_i a_i^2},\\
A_{\max}&=\frac{\max_i a_i}{\sum_i a_i/c}.&&
\end{aligned}
\end{equation}
Here $b$ is the number of active carriers, $b_{\mathrm{eff}}$ is the effective carrier count, and $A_{\max}$ is the maximum load normalized by the mean load. We also report the Gini coefficient of the assignment counts. Metrics are macro-averaged over units; slash-separated span results denote SnapKV/CriticalKV.

Table~\ref{tab:carrier_concentration} shows that span-based retention reduces $b/c$ by 10--24\% and $b_{\mathrm{eff}}/c$ by 39--50\% relative to H2O, while its normalized maximum load is 2.24--3.15$\times$ that of H2O.

For span-based methods, we define the first and last retained tokens of each contiguous span as boundary carriers. Enrichment is the boundary assignment ratio divided by the boundary carrier fraction; the boundary/interior load ratio is computed only for units containing both carrier types.

Across the 512 budget--layer--KV-head units for Llama/\textsc{NarrativeQA}, SnapKV and CriticalKV respectively have lower $b/c$ than H2O in 91.02\% and 80.47\% of units, lower $b_{\mathrm{eff}}/c$ in 91.80\% and 88.87\%, and higher Gini coefficients in 93.36\% and 90.04\% shown in table~\ref{tab:boundary_overload}. Boundary load exceeds interior load in 512/512 SnapKV units and 511/512 CriticalKV units. The pattern is therefore not confined to a small number of heads, layers, or visualization examples.

\begin{figure*}[!p]
  \centering
  \includegraphics[width=\textwidth,height=\textheight,keepaspectratio]{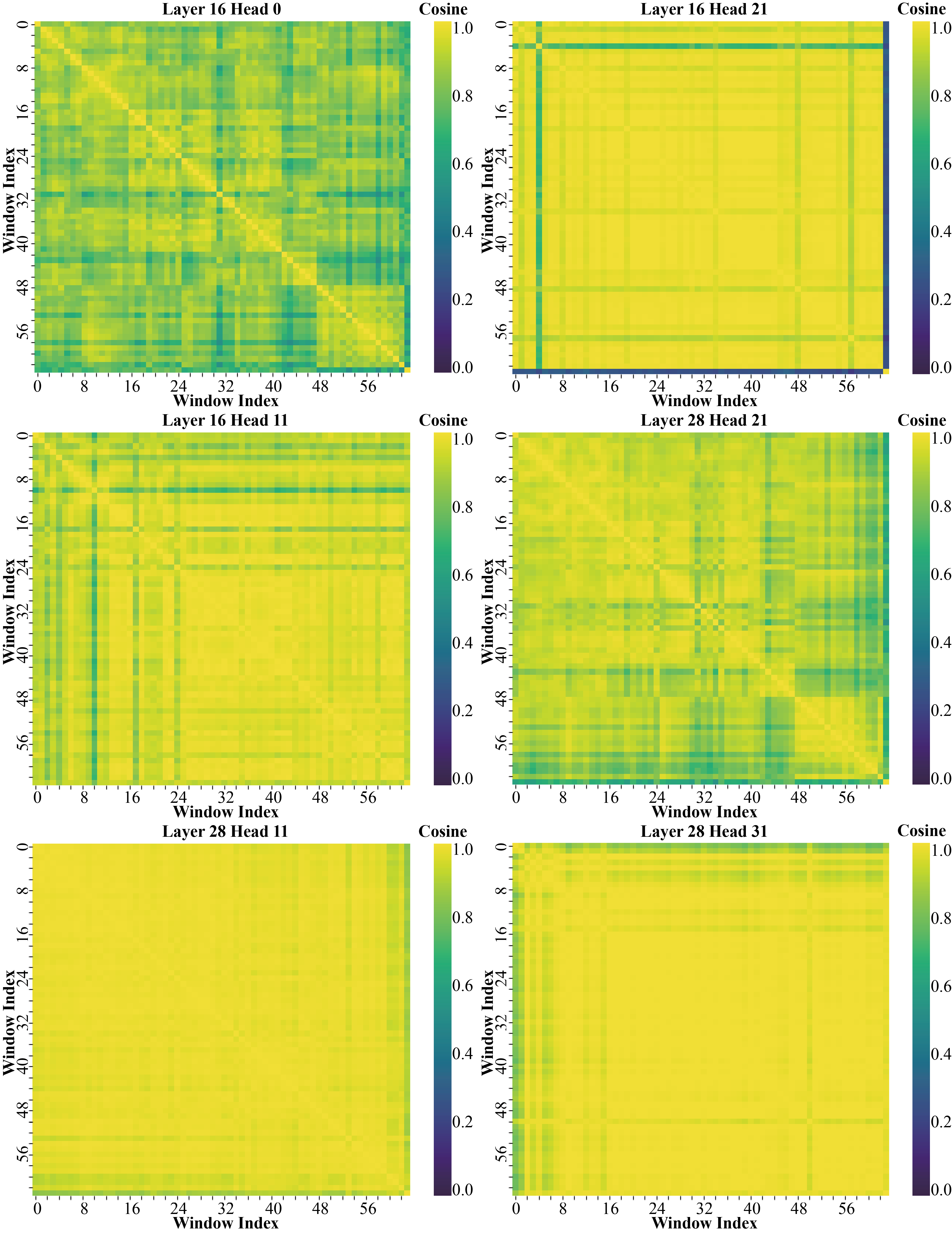}
  \caption{\textbf{Additional cross-window consistency on \textsc{NarrativeQA}.}
  Each heatmap cell reports the cosine similarity between two window summaries $s_W$ computed from full-cache attention outputs at a specific layer and head. We plot several representative heads from middle and later layers. Overall, the matrices are dominated by high-similarity regions (values close to one), indicating that window-level attention outputs are largely consistent across windows from both the prompt (prefill) segment and the held-out future-query segment.}
  \label{fig:window_consistency_narrativeqa}
\end{figure*}

\section{Additional Cross-Window Consistency Visualizations}
\label{appendix:C}

Our method relies on the empirical observation that attention \emph{outputs} are relatively stable across query windows in long contexts, which motivates using a late-prompt (prefill) query window as a surrogate for unseen future queries (Sec.~\ref{sec:query_window}; Fig.~\ref{fig:window_consistency}). We provide additional evidence on \textsc{NarrativeQA}.

\paragraph{Setup.}
For each sequence, we take the first 90\% of tokens as the prompt (prefill segment) and the remaining 10\% of tokens as the future-query segment. We partition the tokens into non-overlapping windows of length $m$, using the same $m$ as in our main analysis. For a fixed layer $\ell$ and head $h$, we compute a window summary by averaging the full-cache attention outputs within each window:
\begin{equation}
  s_W^{(\ell,h)} \;=\; \frac{1}{|W|}\sum_{i\in W} A_{i}^{(\text{full},\ell,h)} V_{\text{full}}^{(\ell,h)} \;\in\; \mathbb{R}^{d_h},
\end{equation}
where $A_{i}^{(\text{full},\ell,h)}$ denotes the full-cache attention weights for query token $i$ at head $(\ell,h)$ and $d_h$ is the head dimension. We then compute cosine similarities between window summaries, forming a cross-window similarity matrix whose $(u,v)$ entry is $\cos\!\big(s_{W_u}^{(\ell,h)}, s_{W_v}^{(\ell,h)}\big)$. In the main text, Fig.~\ref{fig:window_consistency} reports this analysis on \textsc{HotpotQA}; here, we visualize analogous results on \textsc{NarrativeQA}.

\paragraph{Findings.}
Fig.~\ref{fig:window_consistency_narrativeqa} shows that cross-window similarities are typically close to one for most heads, with many matrices appearing almost uniformly bright. This suggests that, across a wide range of layers and heads, the \emph{direction} of the full-cache attention output varies only mildly across query windows, even when comparing windows from different parts of the sequence. We also observe heterogeneity across heads: a minority of heads exhibit more structured variation (e.g., banded patterns or lower similarity involving early windows), indicating modest window-dependent shifts in attention outputs. Nevertheless, the overall high similarity supports the premise underlying GRKV: optimizing the cache to match full-cache outputs on a surrogate window provides a meaningful proxy for the outputs induced by future queries, while avoiding brittle overfitting to any single query token.

\begin{table}[t]
\centering
\small
\setlength{\tabcolsep}{4.0pt}
\renewcommand{\arraystretch}{1.06}
\begin{tabular}{lcc}
\toprule
Surrogate position & LongBench & RULER \\
\midrule
SnapKV (no regression) & 33.96 & 27.44 \\
Early & \textbf{34.76} & 28.50 \\
Middle & 34.34 & 28.33 \\
Late (default) & 34.58 & \textbf{29.09} \\
\midrule
GRKV mean $\pm$ s.d. & $34.56\pm0.21$ & $28.64\pm0.40$ \\
Position variance $s^2$ & 0.044 & 0.159 \\
\bottomrule
\end{tabular}
\caption{Downstream robustness to the surrogate-window position.}
\label{tab:surrogate_position}
\end{table}

\subsection{Surrogate-Window Robustness and Failure Modes}
\label{sec:surrogate_robustness}

\paragraph{Position robustness.}
We evaluate contiguous early, middle, and late prompt windows of $m=32$ using \textit{Llama-3.1-8B-Instruct}, SnapKV, and a 10\% cache budget, with all other settings fixed. Table~\ref{tab:surrogate_position} reports downstream scores rather than attention-output similarity alone.

All positions improve over SnapKV in this ablation, although the late window is best on RULER. This supports empirical positional robustness in the tested setting, not strict position invariance.

\paragraph{Retrieval-oriented stress test.}
We additionally extract the first 32 held-out decoding queries from the Full Cache trajectory on 16K RULER and replay the SnapKV and GRKV caches while fixing the queries and generated prefix. These held-out queries are not used by GRKV. A failure is a sample--head case in which GRKV lowers the surrogate-window reconstruction error but raises the held-out actual-query error.

\begin{table}[t]
\centering
\small
\setlength{\tabcolsep}{4.5pt}
\renewcommand{\arraystretch}{1.06}
\begin{adjustbox}{max width=\columnwidth}
\begin{tabular}{lrrrr}
\toprule
Scope & SnapKV & GRKV & Actual-Q $\Delta$ error & Failure \\
\midrule
All 13 tasks & 27.44 & 29.09 & $-0.00767$ & 2.38\% \\
Eight NIAH tasks & 24.46 & 25.98 & $-0.00595$ & 0.75\% \\
\bottomrule
\end{tabular}
\end{adjustbox}
\caption{Held-out-query retrieval stress test. Negative error changes favor GRKV.}
\label{tab:surrogate_failure}
\end{table}

GRKV improves six NIAH tasks and ties two, but decreases CWE, QA-1, and QA-2 by 1.46, 0.80, and 0.20 points, respectively, showing that gains are not uniform. Failure cases concentrate in the weak-alignment tail shown in table~\ref{tab:surrogate_failure}: the lowest 10\% of head-wise surrogate--actual similarities have a 7.69\% failure rate, compared with 1.79\% for the remaining 90\%. Thus, surrogate optimization transfers on average in this stress test but can fail when surrogate and realized queries are poorly aligned.

\begin{algorithm}[!t]
\caption{Global Regression for KV Cache Merging (GRKV)}
\label{alg:GRKV}
\begin{algorithmic}[1]
\STATE \textbf{Input:} Full cache $K_{\text{full}},V_{\text{full}}$; initial retained cache $K_{\text{Ret}}^0,V_{\text{Ret}}^0$; surrogate queries $Q_{\text{win}}$.
\STATE \textbf{Hyperparameters:} Fixed-token set $\mathcal{F}$; ridge coefficients $\lambda_k,\lambda_v$; maximum steps $S$.
\STATE \textbf{Output:} Merged cache $K_{\text{Ret}}^*,V_{\text{Ret}}^*$.
\STATE $\mathcal{G}\leftarrow\{1,\ldots,c\}\setminus\mathcal{F}$ \hfill // \textit{free retained tokens}
\STATE $Y \leftarrow \operatorname{softmax}(Q_{\text{win}}K_{\text{full}}^\top/\sqrt{d})V_{\text{full}}$
\STATE $K_{\text{Ret}}^{(0)}\leftarrow K_{\text{Ret}}^0$, $V_{\text{Ret}}^{(0)}\leftarrow V_{\text{Ret}}^0$
\FOR{$t=0,\ldots,S-1$}
\STATE $K_{\mathrm{old}}\leftarrow K_{\text{Ret}}^{(t)}$, \quad $V_{\mathrm{old}}\leftarrow V_{\text{Ret}}^{(t)}$
\STATE $X \leftarrow \operatorname{softmax}(Q_{\text{win}}(K_{\text{Ret}}^{(t)})^\top/\sqrt{d})$
\STATE $\begin{aligned}
R_V \leftarrow{} Y-X_{\mathcal{F}}(V_{\text{Ret}}^0)_{\mathcal{F}}-X_{\mathcal{G}}(V_{\text{Ret}}^0)_{\mathcal{G}}
\end{aligned}$
\STATE $Z \leftarrow \big(X_{\mathcal{G}}X_{\mathcal{G}}^\top+\lambda_v I_m\big)^{-1}R_V$
\STATE $(V_{\text{Ret}}^{(t+1)})_{\mathcal{G}}\leftarrow (V_{\text{Ret}}^0)_{\mathcal{G}}+X_{\mathcal{G}}^\top Z$
\STATE $(V_{\text{Ret}}^{(t+1)})_{\mathcal{F}}\leftarrow (V_{\text{Ret}}^0)_{\mathcal{F}}$
\STATE $f(K)=\operatorname{softmax}(Q_{\text{win}}K^\top/\sqrt{d})V_{\text{Ret}}^{(t+1)}$
\STATE $E\leftarrow Y-f(K_{\text{Ret}}^{(t)})$
\STATE $G_{\mathcal{G}}\leftarrow (K_{\text{Ret}}^{(t)})_{\mathcal{G}}-(K_{\text{Ret}}^0)_{\mathcal{G}}$
\STATE $J_{\mathcal{G}} \leftarrow \left.\frac{\partial f}{\partial K_{\mathcal{G}}}\right|_{K=K_{\text{Ret}}^{(t)}}$ \hfill // \textit{vectorized}
\STATE $r_K\leftarrow \operatorname{vec}(E)+J_{\mathcal{G}}\operatorname{vec}(G_{\mathcal{G}})$
\STATE Solve $(J_{\mathcal{G}}J_{\mathcal{G}}^\top+\lambda_k I_{\mathcal{Y}})\alpha = r_K$
\STATE $\operatorname{vec}((K_{\text{Ret}}^{(t+1)})_{\mathcal{G}})\leftarrow \operatorname{vec}((K_{\text{Ret}}^0)_{\mathcal{G}})+J_{\mathcal{G}}^\top\alpha$
\STATE $(K_{\text{Ret}}^{(t+1)})_{\mathcal{F}}\leftarrow (K_{\text{Ret}}^0)_{\mathcal{F}}$
\STATE $\delta_K\leftarrow\|K_{\text{Ret}}^{(t+1)}-K_{\mathrm{old}}\|_{\infty}$
\STATE $\delta_V\leftarrow\|V_{\text{Ret}}^{(t+1)}-V_{\mathrm{old}}\|_{\infty}$
\IF{$\max(\delta_K,\delta_V)\le 1\mathrm{e}{-}9$}
\STATE \textbf{break}
\ENDIF
\ENDFOR
\STATE Let $K_{\text{Ret}}^*,V_{\text{Ret}}^*$ be the final iterates.
\STATE \textbf{return} $K_{\text{Ret}}^*,V_{\text{Ret}}^*$
\end{algorithmic}
\end{algorithm}

\section{Derivations for GRKV Updates}
\label{appendix:D}

This appendix derives the two update rules used by GRKV. We follow the notation in the main text. Define the full-cache pre-projection attention-output target for the surrogate query window $Y \triangleq A_{\text{win}}^{(\text{full})}V_{\text{full}}\in\mathbb{R}^{m\times d}$. The coefficients $\lambda_k>0$ and $\lambda_v>0$ control the strength of regularization for keys and values, respectively. GRKV alternates between a value step, where $K_{\text{Ret}}$ is fixed, and a key step, where $V_{\text{Ret}}$ is fixed.

\subsection{Value-Step Dual Form through the Woodbury Identity}
\label{appendix:value_update}

With the current keys fixed, define
$
X \triangleq A_{\text{win}}^{(\text{Ret})}
=
\operatorname{softmax}\!\left(Q_{\text{win}}K_{\text{Ret}}^\top/\sqrt{d}\right)\in\mathbb{R}^{m\times c}$.
The value-step objective in the main text is:
\begin{equation}
\min_{V_{\text{Ret}}}
\left\|Y-XV_{\text{Ret}}\right\|_F^2
+
\lambda_v\left\|V_{\text{Ret}}-V_{\text{Ret}}^0\right\|_F^2~.
\label{eq:appendixD_value_objective}
\end{equation}

\paragraph{Primal solution.}
Define $\Delta V=V_{\text{Ret}}-V_{\text{Ret}}^0$, $M=\left(X^\top X+\lambda_v I_c\right)$, $N=\left(XX^\top+\lambda_v I_m\right)$, and
$E_V=Y-XV_{\text{Ret}}^0$. Substituting
$V_{\text{Ret}}=V_{\text{Ret}}^0+\Delta V$ into Eq.~\eqref{eq:appendixD_value_objective} gives the ridge problem:
\begin{equation}
\min_{\Delta V}
\left\|E_V-X\Delta V\right\|_F^2
+
\lambda_v\left\|\Delta V\right\|_F^2~.
\label{eq:appendixD_ridge_delta}
\end{equation}
Setting the derivative with respect to $\Delta V$ to zero yields:
\begin{equation}
M\Delta V
=
X^\top E_V~,
\end{equation}
and therefore:
\begin{equation}
\begin{aligned}
V_{\text{Ret}}^*
&= V_{\text{Ret}}^0 + 
M^{-1}
X^\top
\left(Y-XV_{\text{Ret}}^0\right) \\
&=
M^{-1}
\left(X^\top Y+\lambda_v V_{\text{Ret}}^0\right),
\end{aligned}
\label{eq:appendixD_value_primal}
\end{equation}
which is the closed form in Eq.~\ref{eq:eq7}. Here, $V_{\text{Ret}}^*$ corresponds to the merged value cache.

\paragraph{Dual form.}
When $c$ is large, solving the $c\times c$ system can be expensive. Using the ridge identity:
\begin{equation}
M^{-1} X^\top
=
X^\top N^{-1},
\label{eq:appendixD_value_identity}
\end{equation}
Eq.~\eqref{eq:appendixD_value_primal} becomes:
\begin{equation}
V_{\text{Ret}}^*
=
V_{\text{Ret}}^0
+
X^\top
N^{-1}
\left(Y-XV_{\text{Ret}}^0\right).
\end{equation}
Equivalently, define the window-sized variable $
Z
\triangleq
N^{-1}
\left(Y-XV_{\text{Ret}}^0\right)
\in\mathbb{R}^{m\times d}$. Then:
\begin{equation}
V_{\text{Ret}}^*
=
V_{\text{Ret}}^0+X^\top Z~.
\end{equation}
This matches the dual formulation in the main text. The dual form requires solving an $m\times m$ system rather than a $c\times c$ system, which is efficient because GRKV uses a small surrogate window (e.g., $m=32$) while $c$ can be much larger.

\begin{table*}[t]
\centering
\small
\setlength{\tabcolsep}{5pt}
\renewcommand{\arraystretch}{1.05}
\begin{tabular*}{\textwidth}{@{\extracolsep{\fill}}lllrlr@{}}
\toprule
Task & Task Type & Eval. Metric & Avg. Len. & Language & Sample Count \\
\midrule
NarrativeQA         & Single-Doc. QA      & F1       & 18,409 & EN             & 200 \\
Qasper              & Single-Doc. QA      & F1       & 3,619  & EN             & 200 \\
MultiFieldQA-en     & Single-Doc. QA      & F1       & 4,559  & EN             & 150 \\
HotpotQA            & Multi-Doc. QA       & F1       & 9,151  & EN             & 200 \\
2WikiMultihopQA     & Multi-Doc. QA       & F1       & 4,887  & EN             & 200 \\
MuSiQue             & Multi-Doc. QA       & F1       & 11,214 & EN             & 200 \\
GovReport           & Summarization       & ROUGE-L  & 8,734  & EN             & 200 \\
QMSum               & Summarization       & ROUGE-L  & 10,614 & EN             & 200 \\
MultiNews           & Summarization       & ROUGE-L  & 2,113  & EN             & 200 \\
TREC                & Few-shot Learning   & Accuracy & 5,177  & EN             & 200 \\
TriviaQA            & Few-shot Learning   & F1       & 8,209  & EN             & 200 \\
SAMSum              & Few-shot Learning   & ROUGE-L  & 6,258  & EN             & 200 \\
PassageCount        & Synthetic           & Accuracy & 11,141 & EN             & 200 \\
PassageRetrieval-en & Synthetic           & Accuracy & 9,289  & EN             & 200 \\
LCC                 & Code                & Edit Sim & 1,235  & Python/C\#/Java & 500 \\
RepoBench-P         & Code                & Edit Sim & 4,206  & Python/Java     & 500 \\
\bottomrule
\end{tabular*}
\caption{Details of the 16 LongBench datasets.}
\label{tab:longbench_details}
\end{table*}

\begin{table*}[t]
\centering
\small
\setlength{\tabcolsep}{5pt}
\renewcommand{\arraystretch}{1.05}
\begin{tabular*}{\textwidth}{@{\extracolsep{\fill}}lllrlr@{}}
\toprule
Task & Task Type & Eval. Metric & Avg. Len. & Language & Sample Count \\
\midrule
NIAH-S1         & Retrieval / Passkey       & String Match & 16,384 & EN & 500 \\
NIAH-S2         & Retrieval / Vanilla NIAH   & String Match & 16,384 & EN & 500 \\
NIAH-S3         & Retrieval / Long Value     & String Match & 16,384 & EN & 500 \\
NIAH-MK1        & Retrieval / Multi-Key      & String Match & 16,384 & EN & 500 \\
NIAH-MK2        & Retrieval / Line Retrieval & String Match & 16,384 & EN & 500 \\
NIAH-MK3        & Retrieval / KV Retrieval   & String Match & 16,384 & EN & 500 \\
NIAH-MV         & Retrieval / Multi-Value    & String Match & 16,384 & EN & 500 \\
NIAH-MQ         & Retrieval / Multi-Query    & String Match & 16,384 & EN & 500 \\
VT              & Multi-Hop Tracing          & String Match & 16,384 & EN & 500 \\
CWE             & Aggregation                & String Match & 16,384 & EN & 500 \\
FWE             & Aggregation                & String Match & 16,384 & EN & 500 \\
QA-1 (SQuAD)    & Question Answering         & String Match & 16,384 & EN & 500 \\
QA-2 (HotpotQA) & Question Answering         & String Match & 16,384 & EN & 500 \\
\bottomrule
\end{tabular*}
\caption{Details of the 13 RULER tasks at the 16K context-length setting.}
\label{tab:ruler_details}
\end{table*}

\begin{table*}[t]
\centering
\small
\setlength{\tabcolsep}{2.4pt}
\renewcommand{\arraystretch}{1.08}

\begin{adjustbox}{max width=\textwidth}
\begin{tabular}{l*{18}{c}}
\toprule
\multirow{2}{*}{Method}
& \multicolumn{3}{c}{Single-Document QA}
& \multicolumn{3}{c}{Multi-Document QA}
& \multicolumn{3}{c}{Summarization}
& \multicolumn{3}{c}{Few-shot Learning}
& \multicolumn{2}{c}{Synthetic}
& \multicolumn{2}{c}{Code}
& \multirow{2}{*}{Avg.}
& \multirow{2}{*}{Better} \\
\cmidrule(lr){2-4}\cmidrule(lr){5-7}\cmidrule(lr){8-10}\cmidrule(lr){11-13}\cmidrule(lr){14-15}\cmidrule(lr){16-17}
& \rotatebox[origin=c]{45}{NtrvQA}
& \rotatebox[origin=c]{45}{Qasper}
& \rotatebox[origin=c]{45}{MF-en}
& \rotatebox[origin=c]{45}{Hotpot}
& \rotatebox[origin=c]{45}{2WikiQA}
& \rotatebox[origin=c]{45}{Musique}
& \rotatebox[origin=c]{45}{GovRep}
& \rotatebox[origin=c]{45}{QMSum}
& \rotatebox[origin=c]{45}{MultiNews}
& \rotatebox[origin=c]{45}{TREC}
& \rotatebox[origin=c]{45}{TriviaQA}
& \rotatebox[origin=c]{45}{SAMSum}
& \rotatebox[origin=c]{45}{PCount}
& \rotatebox[origin=c]{45}{PR-en}
& \rotatebox[origin=c]{45}{Lcc}
& \rotatebox[origin=c]{45}{RB-P} \\
\midrule

\multicolumn{19}{c}{Llama-3.1-8B-Instruct, 20\% Cache Budget} \\
\midrule
Full Cache  & 30.46 & 47.29 & 55.03 & 58.65 & 51.61 & 33.03 & 35.03 & 24.97 & 27.04 & 29.00 & 84.81 & 39.29 & 10.15 & 100.00 & 51.43 & 44.19 & 45.12 & -- \\
\midrule
SnapKV      & 27.84 & 27.19 & 34.18 & 54.98 & 37.90 & 24.48 & 27.76 & 21.86 & 22.72 & 38.00 & 83.01 & 40.94 & 7.55 & 89.50 & 51.73 & 44.31 & 39.62 & 0/16 \\
w. CaM      & 26.16 & 26.61 & 35.54 & 54.42 & 37.46 & 26.00 & 27.03 & 21.87 & 22.59 & 20.07 & 84.78 & 39.56 & 8.68 & 86.50 & 46.53 & 42.72 & 37.91 & 5/16 \\
w. D2O      & 25.11 & 19.71 & 29.16 & 48.41 & 25.52 & 19.83 & 24.91 & 19.93 & 20.88 & 33.00 & 82.41 & 38.91 & 7.04 & 66.50 & 50.08 & 45.05 & 34.78 & 1/16 \\
w. AsymKV   & 25.60 & 30.06 & 36.48 & 49.08 & 33.52 & 19.46 & 27.92 & 22.29 & 24.40 & 17.50 & 89.24 & 26.99 & 5.61 & 63.50 & 49.10 & 44.95 & 35.36 & 7/16 \\
\rowcolor[RGB]{225,244,252}
w. GRKV     & 27.73 & 28.30 & 34.36 & 54.45 & 36.85 & 25.99 & 27.98 & 22.04 & 23.11 & 41.50 & 84.06 & 41.17 & 8.55 & 92.00 & 51.47 & 44.86 & \textbf{40.28} & \textbf{12/16} \\
\midrule
CriticalKV  & 28.94 & 31.35 & 39.22 & 53.57 & 41.79 & 25.94 & 29.27 & 23.03 & 23.40 & 49.00 & 80.58 & 41.01 & 9.55 & 96.50 & 52.89 & 45.99 & 42.00 & 0/16 \\
w. CaM      & 29.29 & 30.05 & 39.26 & 53.79 & 39.00 & 27.60 & 28.51 & 22.76 & 23.42 & 22.00 & 82.55 & 39.45 & 8.79 & 96.50 & 46.94 & 44.19 & 39.63 & 6/16 \\
w. D2O      & 27.18 & 26.55 & 31.28 & 49.36 & 31.41 & 22.61 & 25.97 & 21.18 & 21.95 & 36.00 & 82.81 & 39.93 & 10.00 & 74.50 & 50.85 & 45.46 & 37.32 & 2/16 \\
w. AsymKV   & 22.76 & 29.76 & 34.17 & 43.08 & 34.27 & 16.60 & 27.21 & 22.12 & 24.24 & 7.00 & 83.87 & 21.86 & 3.83 & 45.50 & 51.92 & 47.17 & 32.21 & 3/16 \\
\rowcolor[RGB]{225,244,252}
w. GRKV     & 29.24 & 32.87 & 39.86 & 53.47 & 42.68 & 26.13 & 29.27 & 23.04 & 23.56 & 50.00 & 81.83 & 41.68 & 10.05 & 97.00 & 53.55 & 46.15 & \textbf{42.52} & \textbf{14/16} \\
\midrule

\multicolumn{19}{c}{Mistral-7B-Instruct-v0.3, 20\% Cache Budget} \\
\midrule
Full Cache  & 28.38 & 40.72 & 51.68 & 49.00 & 36.21 & 27.81 & 34.23 & 25.41 & 26.93 & 55.75 & 84.93 & 20.94 & 5.05 & 98.00 & 46.92 & 53.86 & 42.86 & -- \\
\midrule
SnapKV      & 18.06 & 18.46 & 36.68 & 39.70 & 25.67 & 20.95 & 27.83 & 21.68 & 22.62 & 43.75 & 86.42 & 23.42 & 4.55 & 89.50 & 51.47 & 52.60 & 36.46 & 0/16 \\
w. CaM      & 19.50 & 18.30 & 35.13 & 40.35 & 21.94 & 20.41 & 27.57 & 21.88 & 22.81 & 36.50 & 84.68 & 14.04 & 2.47 & 87.17 & 47.69 & 50.85 & 34.46 & 4/16 \\
w. D2O      & 16.28 & 10.77 & 34.09 & 35.27 & 22.61 & 14.74 & 25.59 & 19.93 & 20.92 & 35.75 & 86.47 & 28.09 & 5.26 & 55.00 & 50.65 & 52.16 & 32.10 & 3/16 \\
w. AsymKV   & 24.57 & 24.10 & 35.83 & 42.84 & 28.98 & 16.77 & 28.72 & 21.64 & 23.78 & 45.82 & 85.61 & 24.71 & 1.61 & 60.17 & 45.81 & 48.68 & 34.98 & 8/16 \\
\rowcolor[RGB]{225,244,252}
w. GRKV     & 18.70 & 19.45 & 37.81 & 39.50 & 27.82 & 21.44 & 28.03 & 22.15 & 23.00 & 44.00 & 87.17 & 23.13 & 4.65 & 90.00 & 52.51 & 52.54 & \textbf{36.99} & \textbf{13/16} \\
\midrule
CriticalKV  & 22.88 & 25.68 & 42.55 & 42.53 & 31.99 & 18.68 & 28.76 & 23.02 & 23.22 & 47.25 & 86.99 & 23.14 & 4.65 & 93.00 & 49.49 & 51.65 & 38.47 & 0/16 \\
w. CaM      & 22.93 & 24.64 & 41.20 & 45.48 & 33.70 & 22.26 & 28.27 & 22.86 & 23.42 & 35.07 & 85.44 & 11.63 & 4.25 & 90.00 & 48.95 & 50.17 & 36.89 & 5/16 \\
w. D2O      & 19.28 & 15.96 & 35.13 & 36.64 & 25.84 & 16.48 & 26.28 & 21.13 & 21.76 & 41.50 & 87.08 & 26.98 & 4.43 & 57.50 & 50.61 & 51.26 & 33.62 & 3/16 \\
w. AsymKV   & 23.06 & 24.27 & 37.25 & 45.63 & 26.43 & 18.94 & 28.80 & 21.58 & 24.07 & 51.00 & 85.36 & 23.25 & 2.43 & 54.50 & 45.06 & 49.23 & 35.05 & 7/16 \\
\rowcolor[RGB]{225,244,252}
w. GRKV     & 23.20 & 25.84 & 43.02 & 44.07 & 32.92 & 19.64 & 29.07 & 23.08 & 23.69 & 47.00 & 88.03 & 24.30 & 4.88 & 93.50 & 50.49 & 52.02 & \textbf{39.05} & \textbf{15/16} \\
\bottomrule
\end{tabular}
\end{adjustbox}
\caption{Detailed scores on all 16 LongBench tasks. The Avg. column reports the average score over all tasks, and the Better column reports the number of tasks on which a method outperforms its base eviction method.}
\label{tab:longbench_detailed_20}
\end{table*}

\subsection{Local Dual Key Update}
\label{appendix:key_update}

With the current values $V_{\text{Ret}}$ fixed, define $f(K)=\operatorname{softmax}\!\left(Q_{\text{win}}K^\top/\sqrt{d}\right)V_{\text{Ret}}$.
At the current key cache $K_{\text{Ret}}$, let $E=Y-f(K_{\text{Ret}})$, $G=K_{\text{Ret}}-K_{\text{Ret}}^0$.
The key objective is nonlinear because $K_{\text{Ret}}$ appears inside the softmax. For an update $\Delta K$, we use the first-order approximation:
\begin{equation}
f(K_{\text{Ret}}+\Delta K)
\approx
f(K_{\text{Ret}})
+
\mathcal{J}[\Delta K],
\end{equation}
where $\mathcal{J}$ is the derivative operator of $f$ with respect to $K$, evaluated at the current $K_{\text{Ret}}$. Let $J\in\mathbb{R}^{md\times cd}$ denote the vectorized matrix representation of $\mathcal{J}$, and let $e=\operatorname{vec}(E)$, $\delta=\operatorname{vec}(\Delta K)$, and $g=\operatorname{vec}(G)$. Substituting the linearization into the key objective gives the local ridge problem:
\begin{equation}
\min_{\delta}
\left\|e-J\delta\right\|_2^2
+
\lambda_k\left\|\delta+g\right\|_2^2~.
\label{eq:key_local_ridge}
\end{equation}
Let $z=\delta+g$. Eq.~\eqref{eq:key_local_ridge} becomes:
\begin{equation}
\min_z
\left\|e+Jg-Jz\right\|_2^2
+
\lambda_k\left\|z\right\|_2^2~.
\end{equation}
The normal equation is:
\begin{equation}
\left(J^\top J+\lambda_k I_{cd}\right)z
=
J^\top(e+Jg)~.
\end{equation}
Using the standard ridge identity:
\begin{equation}
\left(J^\top J+\lambda_k I_{cd}\right)^{-1}J^\top
=
J^\top\left(JJ^\top+\lambda_k I_{\mathcal{Y}}\right)^{-1},
\end{equation}
where $I_{\mathcal{Y}}\in\mathbb{R}^{md\times md}$ is the identity matrix over the vectorized output space, we obtain:
\begin{equation}
z=J^\top\alpha,
\end{equation}
where $\alpha=\left(JJ^\top+\lambda_k I_{\mathcal{Y}}\right)^{-1}(e+Jg)$. Let $K_{\text{Ret}}^*$ denote the merged key cache. Since $z=\delta+g=\operatorname{vec}(K_{\text{Ret}}+\Delta K-K_{\text{Ret}}^0)$, the updated key cache satisfies the rule:
\begin{equation}
\operatorname{vec}(K_{\text{Ret}}^*)
=
\operatorname{vec}(K_{\text{Ret}}^0)
+
J^\top\alpha~.
\end{equation}
Equivalently, with vectorization implicit:
\begin{equation}
K_{\text{Ret}}^*
=
K_{\text{Ret}}^0
+
\operatorname{unvec}\!\left(J^{\top}\alpha\right).
\end{equation}
This is the form used in Eq.~\ref{eq:key_dual_update}. In implementation, GRKV applies the operator $JJ^\top+\lambda_k I_{\mathcal{Y}}$ in a matrix-free manner using Jacobian-vector and vector-Jacobian products, and solves the dual system with conjugate gradients. This avoids explicitly constructing $J$ or $JJ^\top$.

\subsection{Fixed Tokens during Optimization}
\label{appendix:fix_token}

For clarity, the derivations above assume that all retained tokens are allowed to be updated. When a subset of tokens is fixed, the same derivations apply after restricting the optimization variables to the free tokens while keeping the fixed tokens unchanged. Let $\mathcal{F}$ and $\mathcal{G}$ denote the fixed and free token sets, respectively. For the value step, we first subtract the fixed-token contribution from the target, $Y_{\mathcal{G}} = Y - X_{\mathcal{F}}(V_{\text{Ret}}^0)_{\mathcal{F}}$, and then apply the ridge solve only to $X_{\mathcal{G}}$ and $(V_{\text{Ret}})_{\mathcal{G}}$. For the key step, we similarly restrict the Jacobian $J$ to the columns corresponding to the free key variables.

Algorithm~\ref{alg:GRKV} provides detailed pseudocode.

\section{Benchmark Details for LongBench and RULER}
\label{appendix:E}

We use LongBench (licensed under MIT) and RULER (licensed under the Apache License, Version 2.0) as benchmarks. Our use of these benchmarks is consistent with their intended use. Table~\ref{tab:longbench_details} provides detailed information about the 16 datasets in LongBench. Table~\ref{tab:ruler_details} provides detailed information about the 13 tasks in RULER.

\section{Additional Results on LongBench and RULER at 20\% Cache Budget}
\label{appendix:F}

\begin{table*}[t]
\centering
\small
\setlength{\tabcolsep}{3.0pt}
\renewcommand{\arraystretch}{1.08}

\begin{adjustbox}{max width=\textwidth}
\begin{tabular}{lccccccccccccccc}
\toprule
Method
& \rotatebox[origin=c]{45}{cwe}
& \rotatebox[origin=c]{45}{fwe}
& \rotatebox[origin=c]{45}{niah\_mk1}
& \rotatebox[origin=c]{45}{niah\_mk2}
& \rotatebox[origin=c]{45}{niah\_mk3}
& \rotatebox[origin=c]{45}{niah\_mq}
& \rotatebox[origin=c]{45}{niah\_mv}
& \rotatebox[origin=c]{45}{niah\_s1}
& \rotatebox[origin=c]{45}{niah\_s2}
& \rotatebox[origin=c]{45}{niah\_s3}
& \rotatebox[origin=c]{45}{qa\_1}
& \rotatebox[origin=c]{45}{qa\_2}
& \rotatebox[origin=c]{45}{vt}
& \rotatebox[origin=c]{45}{Avg.}
& \rotatebox[origin=c]{45}{Better} \\
\midrule

\multicolumn{16}{c}{Llama-3.1-8B-Instruct, 16K RULER, 20\% Cache Budget} \\
\midrule
Full Cache & 89.28 & 90.33 & 99.60 & 100.00 & 99.00 & 98.90 & 98.90 & 100.00 & 100.00 & 100.00 & 80.80 & 57.20 & 99.72 & 93.36 & -- \\
\midrule
SnapKV     & 22.68 & 72.47 & 59.00 & 12.00 & 4.60 & 43.20 & 38.85 & 89.80 & 87.20 & 3.20 & 28.60 & 31.80 & 76.08 & 43.81 & 0/13 \\
w. CaM     & 3.58 & 83.80 & 45.20 & 9.00 & 4.20 & 31.90 & 29.20 & 66.80 & 76.20 & 2.80 & 29.00 & 32.80 & 27.80 & 34.02 & 3/13 \\
w. D2O     & 3.90 & 68.53 & 50.80 & 4.00 & 1.60 & 36.35 & 29.30 & 91.40 & 81.20 & 3.20 & 21.80 & 29.00 & 67.64 & 37.59 & 1/13 \\
\rowcolor[RGB]{225,244,252}
w. GRKV    & 20.30 & 75.07 & 65.00 & 14.20 & 4.80 & 46.65 & 41.65 & 91.20 & 91.00 & 3.00 & 28.60 & 32.20 & 77.48 & \textbf{45.47} & \textbf{10/13} \\
\midrule
CriticalKV & 41.90 & 77.07 & 90.40 & 14.60 & 3.60 & 90.30 & 80.90 & 95.00 & 99.00 & 5.80 & 37.00 & 36.60 & 84.88 & 58.23 & 0/13 \\
w. CaM     & 4.94 & 84.00 & 80.60 & 8.40 & 3.20 & 73.00 & 66.85 & 88.60 & 93.80 & 4.60 & 37.40 & 33.60 & 38.16 & 47.47 & 2/13 \\
w. D2O     & 12.90 & 70.13 & 86.40 & 5.40 & 1.60 & 86.90 & 75.55 & 94.20 & 96.40 & 3.80 & 23.80 & 31.00 & 80.04 & 51.39 & 0/13 \\
\rowcolor[RGB]{225,244,252}
w. GRKV    & 42.22 & 78.80 & 91.20 & 15.20 & 3.60 & 90.15 & 81.25 & 95.20 & 99.00 & 6.20 & 37.60 & 37.40 & 85.80 & \textbf{58.74} & \textbf{10/13} \\
\midrule

\multicolumn{16}{c}{Mistral-7B-Instruct-v0.3, 16K RULER, 20\% Cache Budget} \\
\midrule
Full Cache & 82.64 & 88.07 & 97.60 & 95.00 & 77.60 & 88.50 & 88.65 & 94.40 & 96.40 & 99.60 & 72.00 & 49.80 & 96.04 & 86.64 & -- \\
\midrule
SnapKV     & 68.32 & 84.67 & 14.40 & 5.60 & 1.20 & 10.75 & 9.35 & 46.00 & 11.00 & 2.40 & 28.40 & 34.60 & 15.80 & 25.58 & 0/13 \\
w. CaM     & 42.42 & 91.40 & 14.00 & 4.40 & 1.40 & 9.60 & 9.05 & 22.60 & 10.80 & 2.40 & 30.40 & 32.60 & 6.68 & 21.37 & 3/13 \\
w. D2O     & 59.96 & 83.53 & 12.40 & 3.20 & 0.40 & 10.35 & 9.65 & 45.80 & 10.80 & 2.40 & 26.00 & 31.60 & 7.64 & 23.36 & 1/13 \\
\rowcolor[RGB]{225,244,252}
w. GRKV    & 70.38 & 85.00 & 16.60 & 5.60 & 1.40 & 11.35 & 9.80 & 45.40 & 14.60 & 2.40 & 28.60 & 33.60 & 16.28 & \textbf{26.23} & \textbf{9/13} \\
\midrule
CriticalKV & 77.66 & 81.33 & 25.20 & 6.20 & 1.40 & 18.90 & 17.80 & 44.00 & 35.80 & 2.40 & 35.60 & 37.20 & 41.76 & 32.71 & 0/13 \\
w. CaM     & 43.30 & 88.73 & 23.80 & 4.60 & 1.00 & 15.75 & 14.60 & 43.60 & 29.40 & 2.40 & 36.00 & 32.20 & 20.40 & 27.37 & 2/13 \\
w. D2O     & 65.64 & 79.07 & 21.00 & 3.00 & 0.80 & 17.05 & 15.65 & 46.40 & 35.00 & 2.60 & 27.40 & 34.40 & 20.60 & 28.35 & 2/13 \\
\rowcolor[RGB]{225,244,252}
w. GRKV    & 77.40 & 82.13 & 26.60 & 6.00 & 1.60 & 19.50 & 18.20 & 45.40 & 38.00 & 2.40 & 35.60 & 36.80 & 41.88 & \textbf{33.19} & \textbf{8/13} \\
\bottomrule
\end{tabular}
\end{adjustbox}
\caption{Detailed scores on all 13 RULER tasks. The Avg. column reports the average score over all tasks, and the Better column reports the number of tasks on which a method outperforms its base eviction method.}
\label{tab:ruler_detailed_20}
\end{table*}

\begin{table}[!t]
\centering
\small
\setlength{\tabcolsep}{5pt}
\renewcommand{\arraystretch}{1.05}
\newcommand{\groupsep}{\addlinespace[1pt]\specialrule{0.3pt}{1pt}{1pt}}
\begin{tabular*}{\columnwidth}{@{\extracolsep{\fill}}llc@{}}
\toprule
Factor & Setting & Avg. \\
\midrule
Regularization strength
& $\lambda=1$ & 27.95 \\
($\lambda=\lambda_k=\lambda_v$)
& $\lambda=10^{-1}$ & 28.47 \\
& $\lambda=10^{-2}\dagger$ & \textbf{29.09} \\
& $\lambda=0$ & 27.49 \\
\groupsep

Fixed retained-token ratio
& $\beta=0\%$ & 28.48 \\
($\beta$)
& $\beta=10\%\dagger$ & \textbf{29.09} \\
& $\beta=20\%$ & 28.90 \\
& $\beta=30\%$ & 28.94 \\
\groupsep

Update steps
& $S=1\dagger$ & \textbf{29.09} \\
($S$)
& $S=2$ & 28.77 \\
& $S=3$ & 28.84 \\
\groupsep

Surrogate window size
& $m=32\dagger$ & \textbf{29.09} \\
($m$)
& $m=48$ & 27.12 \\
& $m=64$ & 26.50 \\
\groupsep

Sink and window tokens
& Fixed$\dagger$ & \textbf{29.09} \\
& Updated & 28.72 \\
\groupsep

Optimized cache components
& GRK & 29.05 \\
& GRV & 28.40 \\
& GRKV$\dagger$ & \textbf{29.09} \\
\bottomrule
\end{tabular*}
\caption{GRKV ablations on RULER with SnapKV on \textit{Llama-3.1-8B-Instruct} at a 10\% cache budget. Bold indicates the best setting in each group. $\dagger$ denotes the default setting. $\lambda=0$ denotes no regularization.}
\label{tab:ruler_ablation_summary}
\end{table}

\begin{table}[!t]
\centering
\small
\setlength{\tabcolsep}{5pt}
\renewcommand{\arraystretch}{1.08}
\begin{tabular*}{\columnwidth}{@{\extracolsep{\fill}}lllcc@{}}
\toprule
Model & Retention & Method & Base & GRKV \\
\midrule
\multirow{3}{*}{\shortstack[l]{\textit{Llama-3.1}\\\textit{-8B-Instruct}}}
& \multirow{2}{*}{Span-based}
& PyramidKV & 28.55 & \textbf{30.49} \\
& & Ada-KV & 31.50 & \textbf{32.17} \\
\addlinespace[1pt]
\cmidrule{2-5}
\addlinespace[1pt]
& Token-based
& H2O & 19.33 & \textbf{22.43} \\
\midrule
\multirow{2}{*}{\textit{Qwen3-14B}}
& \multirow{2}{*}{Span-based}
& SnapKV & 32.23 & \textbf{34.10} \\
& & CriticalKV & 52.61 & \textbf{53.32} \\
\bottomrule
\end{tabular*}
\caption{RULER compatibility at a 10\% cache budget. Each row compares a base eviction method with its GRKV-enhanced variant.}
\label{tab:compatibility_ruler}
\end{table}

\begin{figure*}[t]
\centering
\includegraphics[width=\textwidth]{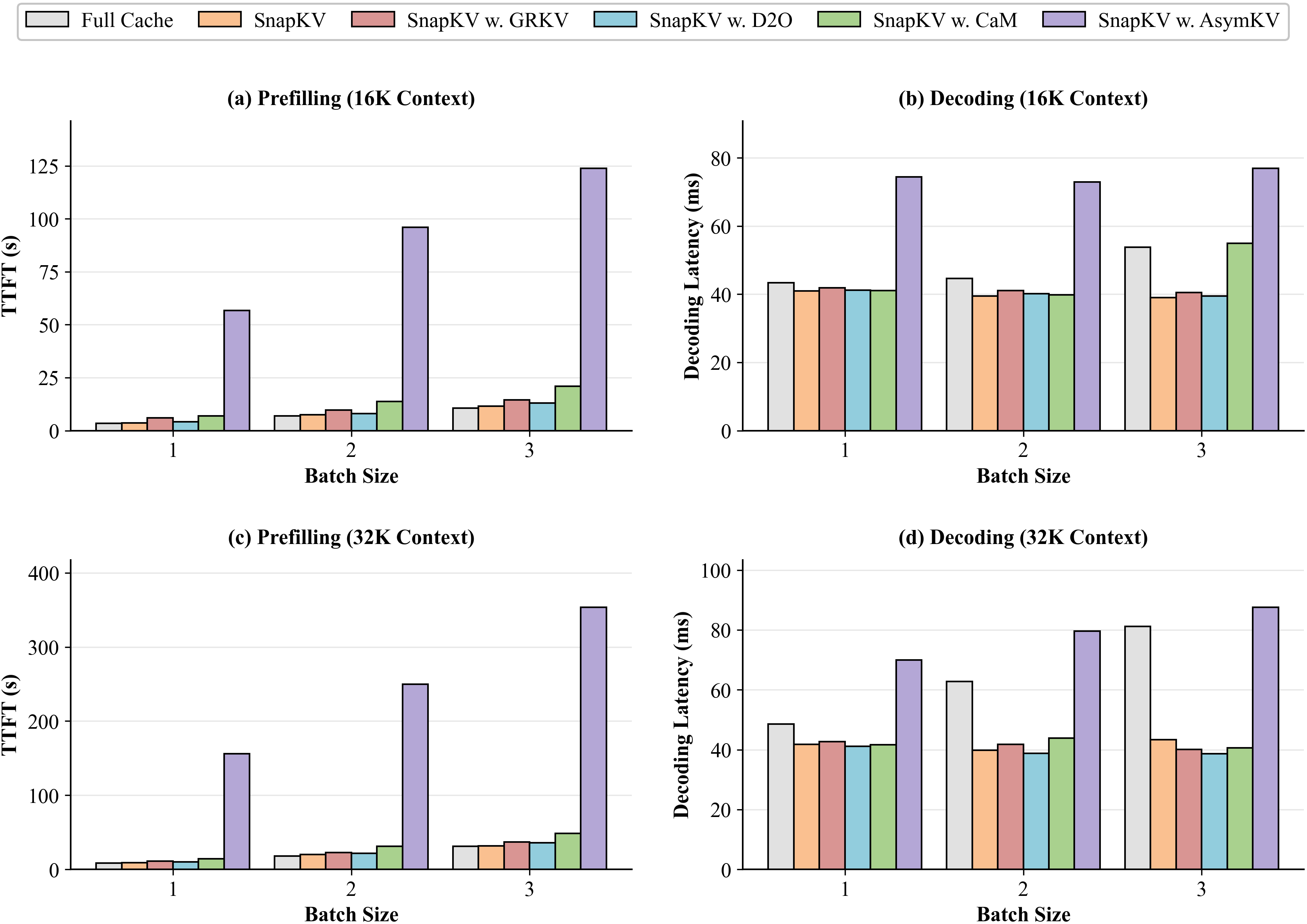}
\caption{TTFT and decoding latency on two A6000 GPUs across 16K--32K context lengths and batch sizes 1--3.}
\label{fig:kv_efficiency_16k_32k}
\end{figure*}

\begin{table*}[t]
\centering
\scriptsize
\setlength{\tabcolsep}{3.8pt}
\renewcommand{\arraystretch}{1.05}
\begin{tabular*}{\textwidth}{@{\extracolsep{\fill}}lccccccc@{}}
\toprule
Configuration & $\lambda$ & $m$ & $S$ & LB loss ($\times10^{-3}$) & LongBench & RULER loss ($\times10^{-3}$) & RULER \\
\midrule
SnapKV & -- & -- & -- & 7.102 & 33.96 & 7.268 & 27.44 \\
$+$ CaM & -- & -- & -- & 9.767 & 32.27 & 10.697 & 18.19 \\
$+$ D2O & -- & -- & -- & 17.036 & 27.44 & 8.784 & 23.35 \\
$+$ AsymKV & -- & -- & -- & 14.726 & 28.97 & -- & -- \\
$+$ GRKV & 1 & 32 & 1 & 6.905 & 34.33 & 6.885 & 27.95 \\
$+$ GRKV & 0.1 & 32 & 1 & 6.799 & 34.29 & 6.463 & 28.47 \\
$+$ GRKV (default) & 0.01 & 32 & 1 & \textbf{5.940} & \textbf{34.58} & \textbf{5.655} & \textbf{29.09} \\
$+$ GRKV & 0 & 32 & 1 & 2.283 & 32.06 & 2.173 & 27.49 \\
$+$ GRKV & 0.01 & 48 & 1 & 7.621 & 33.86 & 7.387 & 27.12 \\
$+$ GRKV & 0.01 & 64 & 1 & 7.838 & 33.49 & 7.616 & 26.50 \\
$+$ GRKV & 0.01 & 32 & 2 & 5.035 & 34.33 & 4.793 & 28.77 \\
$+$ GRKV & 0.01 & 32 & 3 & 4.735 & 34.19 & 4.508 & 28.84 \\
\bottomrule
\end{tabular*}
\caption{Common reconstruction discrepancy and downstream performance across merging methods and GRKV configurations.}
\label{tab:loss_performance}
\end{table*}

\begin{table*}[t]
\centering
\small
\setlength{\tabcolsep}{4.2pt}
\renewcommand{\arraystretch}{1.06}
\begin{tabular*}{\textwidth}{@{\extracolsep{\fill}}llccccc@{}}
\toprule
Model & Base & Base score & $+$GRKV score & Gain & 95\% paired-bootstrap CI & Exact $p$ \\
\midrule
Llama-3.1-8B & SnapKV & $33.95\pm0.09$ & $34.63\pm0.10$ & \textbf{+0.68} & \textbf{[+0.45,+0.94]} & $3.1\times10^{-5}$ \\
Llama-3.1-8B & CriticalKV & $35.89\pm0.15$ & $36.60\pm0.03$ & \textbf{+0.71} & \textbf{[+0.47,+0.98]} & $3.1\times10^{-5}$ \\
Mistral-7B & SnapKV & $33.07\pm0.05$ & $33.85\pm0.09$ & \textbf{+0.78} & \textbf{[+0.35,+1.29]} & $1.9\times10^{-3}$ \\
Mistral-7B & CriticalKV & $33.72\pm0.06$ & $34.35\pm0.10$ & \textbf{+0.63} & \textbf{[+0.36,+0.92]} & $4.9\times10^{-4}$ \\
\bottomrule
\end{tabular*}
\caption{Three-seed LongBench results and task-level paired significance tests.}
\label{tab:multiseed}
\end{table*}

\begin{table}[t]
\centering
\small
\setlength{\tabcolsep}{4.0pt}
\renewcommand{\arraystretch}{1.06}
\begin{adjustbox}{max width=\columnwidth}
\begin{tabular}{rcccc}
\toprule
Context & Full peak & SnapKV peak & GRKV peak & Full $\rightarrow$ GRKV KV \\
\midrule
16K & 18.79 & 17.01 & 17.01 & $2.00\rightarrow0.20$ \\
32K & 22.61 & 19.01 & 19.01 & $4.00\rightarrow0.40$ \\
\bottomrule
\end{tabular}
\end{adjustbox}
\caption{A100 prefill peak and post-prefill KV memory (GiB), with BF16, FlashAttention-2, batch size 1, and a 10\% budget.}
\label{tab:prefill_peak_memory}
\end{table}

This appendix reports detailed per-task results under a larger 20\% KV-cache budget on LongBench and RULER. All settings follow the main text. Compared with the 10\% budget in the main experiments, the 20\% budget preserves more tokens after eviction. In this less aggressive compression regime, GRKV remains the most reliable KV-cache merging method across benchmarks, model backbones, and base eviction methods.

\paragraph{LongBench (20\% budget).}
Table~\ref{tab:longbench_detailed_20} reports per-task scores on all 16 LongBench tasks. GRKV consistently improves the average score of both base eviction methods across both model backbones. On \textit{Llama-3.1-8B-Instruct}, it improves SnapKV from 39.62 to 40.28, with gains on 12/16 tasks, and improves CriticalKV from 42.00 to 42.52, with gains on 14/16 tasks. On \textit{Mistral-7B-Instruct-v0.3}, it raises SnapKV from 36.46 to 36.99 and CriticalKV from 38.47 to 39.05, improving 13/16 and 15/16 tasks, respectively. These gains are more consistent than those of the local KV-cache merging baselines. CaM, D2O, and AsymKV occasionally improve individual tasks, but they reduce the average score in all four LongBench settings. This pattern is consistent with the main-text results: when the retained cache already contains more tokens, local merging can still over-aggregate information, whereas GRKV's global, ridge-regularized reconstruction provides a more stable way to recover information from evicted tokens.

\paragraph{RULER (16K, 20\% budget).}
Table~\ref{tab:ruler_detailed_20} reports results on all 13 RULER tasks. GRKV again improves both base eviction methods across both model backbones. On \textit{Llama-3.1-8B-Instruct}, SnapKV with GRKV improves the average score from 43.81 to 45.47, with gains on 10/13 tasks, and CriticalKV with GRKV improves from 58.23 to 58.74, also with gains on 10/13 tasks. On \textit{Mistral-7B-Instruct-v0.3}, SnapKV with GRKV improves from 25.58 to 26.23, with gains on 9/13 tasks, while CriticalKV with GRKV improves from 32.71 to 33.19, with gains on 8/13 tasks. The improvements are most visible on retrieval-oriented NIAH variants and variable tracking, where useful evidence can be dispersed across the context. In contrast, CaM and D2O often reduce the average score relative to the corresponding base eviction method, especially in retrieval-heavy settings. These 20\% budget results reinforce the main conclusion that GRKV is robust across cache budgets and that global reconstruction is more dependable than local KV-cache merging under span-based retention.

\section{Additional Results on Ablations, Compatibility, and Efficiency}
\label{appendix:G}

Table~\ref{tab:ruler_ablation_summary} reports additional ablations on RULER using SnapKV with GRKV on \textit{Llama-3.1-8B-Instruct} under a 10\% cache budget. The trends are consistent with the LongBench ablations in the main text.

\textbf{Regularization strength.} Regularization remains important: the default $\lambda_k=\lambda_v=10^{-2}$ achieves the best average score (29.09), while removing regularization lowers the score to 27.49. Stronger regularization also weakens performance, with $\lambda_k=\lambda_v=10^{-1}$ and $\lambda_k=\lambda_v=1$ obtaining 28.47 and 27.95, respectively. This supports the role of ridge regularization in allowing useful reconstruction while preventing overly aggressive KV-cache updates.

\textbf{Fixed retained-token ratio.} The fixed retained-token ratio also affects performance. Fixing the top $\beta=10\%$ of high-attention retained tokens gives the best score (29.09), outperforming both updating all retained tokens (28.48) and fixing larger ratios, such as $\beta=20\%$ (28.90) or $\beta=30\%$ (28.94). This suggests that preserving a small set of important anchor tokens improves optimization stability, while fixing too many retained tokens restricts the carrier set available for global reconstruction.

\textbf{Update steps.} The number of alternating update steps follows a similar pattern to the main-text ablation. A single update step performs best (29.09), while increasing the number of steps to $S=2$ or $S=3$ lowers the average score to 28.77 and 28.84. This indicates that additional alternating updates do not improve generalization to downstream queries and may overfit the surrogate window.

\textbf{Surrogate window size.} The surrogate window size has the largest effect in this RULER sweep. The default $m=32$ achieves the highest score (29.09), while increasing the window size to $m=48$ and $m=64$ lowers the average to 27.12 and 26.50, respectively. This supports using $m=32$ in the main experiments, which matches the window size used by the eviction baselines.

\textbf{Fixed sink and window tokens.} Fixing sink and surrogate-window tokens is beneficial: updating them reduces performance from 29.09 to 28.72. 

\textbf{Optimized cache components.} Updating both keys and values performs best overall (29.09), but the key-only variant GRK is very close (29.05), while value-only updating is weaker (28.40). These results suggest that key reconstruction is especially important on RULER, while joint key-value optimization still gives the best default configuration.

\textbf{Compatibility.} Table~\ref{tab:compatibility_ruler} reports additional compatibility results on RULER under a 10\% cache budget. The results show that GRKV remains effective across both span-based and token-based retention methods. On \textit{Llama-3.1-8B-Instruct}, GRKV improves the span-based budget-allocation methods PyramidKV and Ada-KV, which dynamically allocate KV-cache budgets across layers and heads, respectively. Specifically, PyramidKV improves from 28.55 to 30.49, and Ada-KV improves from 31.50 to 32.17. GRKV also improves the token-based H2O baseline from 19.33 to 22.43, indicating that its global reconstruction objective is not limited to span-based retention. On the larger \textit{Qwen3-14B} model, GRKV further improves span-based SnapKV from 32.23 to 34.10 and CriticalKV from 52.61 to 53.32. These results complement the LongBench compatibility results in the main text and show that GRKV consistently improves different eviction backbones across retention granularities, model scales, and benchmarks.

\textbf{Efficiency.} We additionally evaluate efficiency on two A6000 GPUs with batch sizes 1, 2, and 3, using context lengths of 16K and 32K. As shown in Fig.~\ref{fig:kv_efficiency_16k_32k}, SnapKV with GRKV preserves the efficient decoding behavior of SnapKV while adding a moderate prefill cost for the global regression update. At 16K, SnapKV with GRKV increases prefill latency from 3.65 to 5.91~s at batch size 1, from 7.42 to 9.66~s at batch size 2, and from 11.47 to 14.41~s at batch size 3. These costs remain substantially lower than those of AsymKV, which reaches 56.77, 95.93, and 123.82~s, and are also lower than CaM's costs at larger batch sizes. In decoding, SnapKV with GRKV remains close to SnapKV: at 16K, decoding latency is 41.93, 41.13, and 40.49~ms/token for batch sizes 1, 2, and 3, compared with 40.97, 39.47, and 39.04~ms/token for SnapKV.

The same trend holds at 32K. SnapKV with GRKV has moderate prefill overhead relative to SnapKV, increasing TTFT from 9.19 to 11.28~s at batch size 1, from 20.31 to 22.69~s at batch size 2, and from 31.45 to 36.79~s at batch size 3. This overhead is comparable to D2O and much smaller than CaM and AsymKV, with AsymKV reaching 155.87--353.55~s across batch sizes. During decoding, SnapKV with GRKV again stays near SnapKV and other compression methods: SnapKV with GRKV obtains 42.64, 41.76, and 40.03~ms/token across batch sizes, while Full Cache grows from 48.57 to 81.12~ms/token as the batch size increases. These results support the main-text observation that GRKV preserves the decoding benefits of eviction-based compression, while its additional regression step mainly affects prefill and remains practical compared with heavier merging methods.

\begin{table*}[t]
\centering
\scriptsize
\setlength{\tabcolsep}{3.8pt}
\renewcommand{\arraystretch}{1.06}
\begin{tabular*}{\textwidth}{@{\extracolsep{\fill}}lcccc@{}}
\toprule
Method & 16K TTFT (s), BS1/2 & 32K TTFT (s), BS1/2 & 16K decode (ms/token), BS1/2 & 32K decode (ms/token), BS1/2 \\
\midrule
Full Cache & 8.34 / 25.06 & 30.66 / 94.81 & 84.45 / 113.29 & 153.51 / 184.76 \\
SnapKV & 9.01 / 26.19 & 31.79 / 95.54 & 27.76 / 28.24 & 27.74 / 28.38 \\
SnapKV $+$ GRKV & \textbf{9.17 / 26.39} & \textbf{32.61 / 96.92} & \textbf{28.43 / 29.74} & \textbf{28.44 / 29.67} \\
SnapKV $+$ D2O & 9.02 / 26.51 & 32.49 / 97.32 & 27.35 / 28.74 & 27.41 / 28.72 \\
SnapKV $+$ CaM & 20.12 / 48.39 & 50.31 / 133.96 & 28.87 / 29.89 & 28.90 / 29.91 \\
SnapKV $+$ AsymKV & 49.29 / 94.78 & 129.88 / 236.73 & 57.53 / 59.68 & 59.52 / 60.72 \\
\bottomrule
\end{tabular*}
\caption{End-to-end efficiency with a unified eager-attention backend and 10\% cache budget.}
\label{tab:unified_eager_efficiency}
\end{table*}

\begin{table}[t]
\centering
\small
\setlength{\tabcolsep}{3.5pt}
\renewcommand{\arraystretch}{1.06}
\begin{tabular}{rccccc}
\toprule
Budget & Linearize & CG & K update & V update & TTFT \\
\midrule
10\% & 0.02 & 0.31 & 0.33 & 0.06 & 5.48 \\
20\% & 0.04 & 0.57 & 0.61 & 0.11 & 6.07 \\
\bottomrule
\end{tabular}
\caption{GRKV construction-time breakdown in seconds on A100 at 32K.}
\label{tab:construction_breakdown}
\end{table}

\subsection{Loss--Performance Correlation}
\label{sec:loss_performance}

We examine whether the reconstruction objective tracks downstream quality under \textit{Llama-3.1-8B-Instruct}, a 10\% cache budget, and a common SnapKV backbone. For comparability, all configurations use the same 32 diagnostic queries and the normalized discrepancy:
\begin{equation}
\begin{aligned}
\mathcal L_{\mathrm{common}}
&=\frac{\sum_{i,\ell,h}\|Y^{\mathrm{full}}_{i,\ell,h}-Y^{\mathrm{comp}}_{i,\ell,h}\|_F^2}
{\sum_{i,\ell,h}\|Y^{\mathrm{full}}_{i,\ell,h}\|_F^2},\\
Y&=AV.
\end{aligned}
\end{equation}
These diagnostic queries are shared across configurations but are not held-out future queries and are not independent of GRKV optimization.

Across the displayed configurations, LongBench has Pearson $r=-0.836$ ($p=7.02\times10^{-4}$) and Spearman $\rho=-0.592$ ($p=4.26\times10^{-2}$); RULER has Pearson $r=-0.754$ ($p=7.38\times10^{-3}$) and Spearman $\rho=-0.836$ ($p=1.33\times10^{-3}$). Lower discrepancy is therefore associated with higher performance in this comparison, but the $\lambda=0$ and multi-step settings show that the relationship is not strictly monotonic shown in table~\ref{tab:loss_performance}. Regularization and constrained updates remain important for generalization beyond the diagnostic window.

\subsection{Multi-Seed Evaluation and Significance}
\label{sec:multiseed_results}

We repeat the LongBench experiments with seeds 42, 142, and 281. Scores are the mean and sample standard deviation across seeds. For paired inference, we average each task over seeds, bootstrap the 16 task-level paired differences with 100,000 resamples, and enumerate all $2^{16}$ sign assignments for an exact two-sided paired sign-flip test.

GRKV improves all 12/12 seed-level model--base comparisons. All confidence intervals exclude zero and all exact tests have $p<0.002$ shown in table~\ref{tab:multiseed}; this supports stable aggregate gains without implying that every task improves at every seed.

\subsection{Controlled Efficiency, Memory, and Resource Trade-Offs}
\label{sec:controlled_efficiency}

\paragraph{Persistent cache versus transient prefill memory.}
The reported cache budget refers to persistent KV storage after compression. During prefill, GRKV processes layers sequentially: previously processed layers are compressed, while only the current layer retains its full cache and regression workspace before being replaced by the compressed cache.

At the reported precision, GRKV has the same prefill peak as SnapKV and remains 9.5\%/15.9\% below Full Cache at 16K/32K shown in table~\ref{tab:prefill_peak_memory}. This does not mean that prefill never accesses a full layer cache; it distinguishes transient working memory from the 90\% reduction in persistent KV storage.

\paragraph{Unified attention backend.}
To control for FlashAttention compatibility, we run every method with eager attention while retaining each method's original cache-compression implementation. The setup uses one A100, CUDA 12.8, PyTorch 2.9.1, Transformers 4.57.3, CUDA-event timing, one warm-up, and four measured repetitions.

Under this controlled backend, GRKV adds 0.8--2.6\% TTFT over SnapKV and remains close to D2O shown in table~\ref{tab:unified_eager_efficiency}. FlashAttention results in the main text demonstrate compatibility with efficient kernels rather than replacing this controlled comparison.

\begin{table}[t]
\centering
\small
\setlength{\tabcolsep}{4.2pt}
\renewcommand{\arraystretch}{1.06}
\begin{adjustbox}{max width=\columnwidth}
\begin{tabular}{lcccc}
\toprule
Method & Budget & TTFT & Persistent KV & LongBench \\
\midrule
SnapKV & 10\% & 0.57 & 0.080 GiB & 33.96 \\
SnapKV $+$ GRKV & 10\% & 0.69 & 0.080 GiB & 34.58 \\
SnapKV & 12\% & 0.70 & 0.096 GiB & \textbf{34.99} \\
\bottomrule
\end{tabular}
\end{adjustbox}
\caption{Approximate iso-prefill-compute comparison on average-length LongBench inputs.}
\label{tab:iso_prefill}
\end{table}

\begin{table*}[t]
\centering
\scriptsize
\setlength{\tabcolsep}{4.0pt}
\renewcommand{\arraystretch}{1.05}
\begin{tabular*}{\textwidth}{@{\extracolsep{\fill}}clcc@{}}
\toprule
Context & Method & TTFT (s), BS1/2/3 & Decode (ms/token), BS1/2/3 \\
\midrule
16K & Full Cache & 3.17 / 3.40 / 5.13 & 32.58 / 34.59 / 41.78 \\
16K & SnapKV & 3.22 / 3.88 / 5.63 & 29.38 / 29.43 / 31.08 \\
16K & SnapKV $+$ GRKV & \textbf{3.50 / 4.47 / 6.43} & \textbf{30.55 / 30.46 / 30.82} \\
16K & SnapKV $+$ D2O & 3.26 / 4.33 / 6.62 & 30.58 / 29.85 / 31.67 \\
16K & SnapKV $+$ CaM & 10.06 / 20.03 / 30.35 & 30.93 / 29.90 / 30.12 \\
16K & SnapKV $+$ AsymKV$^\dagger$ & 62.92 / 120.99 / 179.06 & 57.71 / 59.87 / 62.02 \\
\midrule
32K & Full Cache & 4.99 / 8.42 / 12.59 & 39.18 / 48.14 / 64.89 \\
32K & SnapKV & 5.50 / 9.31 / 13.67 & 31.62 / 31.59 / 32.45 \\
32K & SnapKV $+$ GRKV & \textbf{6.07 / 10.28 / 14.89} & \textbf{31.19 / 30.34 / 32.37} \\
32K & SnapKV $+$ D2O & 5.68 / 10.34 / 14.20 & 31.56 / 30.95 / 32.07 \\
32K & SnapKV $+$ CaM & 18.14 / 36.47 / 54.16 & 30.90 / 30.01 / 32.03 \\
32K & SnapKV $+$ AsymKV$^\dagger$ & 164.28 / 299.43 / 434.58 & 62.48 / 63.74 / 65.00 \\
\bottomrule
\end{tabular*}
\caption{End-to-end A100 efficiency at a 20\% cache budget. $^\dagger$AsymKV uses eager attention; all other methods use FlashAttention-2.}
\label{tab:efficiency_20_a100}
\end{table*}

\paragraph{Regression overhead and key-update cost.}
Table~\ref{tab:construction_breakdown} isolates the additional construction work at 32K with FlashAttention-2 and batch size 1. The K update equals linearization plus conjugate gradients (CG), including JVP/VJP operations.

The K update contributes 84.6\%/84.7\% of the added regression time at 10\%/20\%, and CG contributes 93--94\% of the K-update time. At 10\%, GRV/GRK/GRKV score 34.40/34.21/34.58 on LongBench and 28.40/29.05/29.09 on RULER. GRV is therefore the lightweight variant, while the key update provides an additional 0.18 points on LongBench and 0.69 on RULER when compute permits.

\paragraph{Iso-memory and approximate iso-prefill comparison.}

At fixed persistent memory, GRKV improves SnapKV by 0.62 points without changing the 0.080-GiB compressed cache. At nearly equal TTFT, however, SnapKV at 12\% is 0.41 points better and uses 20\% more persistent KV memory shown in table~\ref{tab:iso_prefill}. Increasing the cache is therefore preferable when memory permits and prefill latency is the primary constraint; GRKV targets settings where persistent serving memory or concurrency capacity is constrained.

\paragraph{End-to-end results at a 20\% budget.}
Table~\ref{tab:efficiency_20_a100} reports A100 measurements across context lengths and batch sizes; TTFT includes the complete regression solve. Across all six settings, adding GRKV to SnapKV increases TTFT by only 8.7--15.2\%, while changing decoding latency by at most 4.0\%. GRKV also remains competitive with D2O: their TTFT and decoding latency differ by at most 7.4\% and 2.7\%, respectively. Compared with CaM, GRKV reduces TTFT by 65.2--78.8\% while keeping decoding latency within 2.3\%. Relative to Full Cache, GRKV incurs a 10.4--31.5\% one-time TTFT overhead but lowers per-token decoding latency by 6.2--50.1\%, with the reduction becoming particularly pronounced at 32K contexts and larger batch sizes. These results expose the intended serving trade-off: the regression cost is paid once during prefill, whereas the decoding benefit accumulates over generated tokens. We exclude AsymKV from direct timing comparisons because it uses eager attention rather than FlashAttention-2.

\end{document}